\documentclass[journal]{IEEEtran}

\usepackage{amsmath,amsfonts,amssymb}
\usepackage{algorithmic}
\usepackage{algorithm}
\usepackage{array}
\usepackage[caption=false,font=normalsize,labelfont=sf,textfont=sf]{subfig}
\usepackage{textcomp}
\usepackage{stfloats}
\usepackage{url}
\usepackage{verbatim}
\usepackage{graphicx}
\usepackage{cite}
\usepackage{mathrsfs}
\usepackage{circledsteps}
\usepackage{enumerate}
\usepackage{booktabs}
\usepackage{multirow}
\usepackage{makecell}
\usepackage{threeparttable}
\usepackage{bm}

\title{FLOAT Drone for Physical Interaction: Lateral Airflow Reduction, Wrench Modeling, and Adaptive Control}

\author{Junxiao Lin$^*$, Kehan Zhou$^*$, Shuhang Ji, Yimin Peng, Shen Wang, Jialiang Hou$^\dagger$, and Fei Gao$^\dagger$%
\thanks{$^*$Equal contribution.}
\thanks{$^\dagger$Corresponding authors: Jialiang Hou and Fei Gao.}
\thanks{All authors are with the Institute of Cyber-Systems and Control, College of Control Science and Engineering, Zhejiang University, Hangzhou 310027, China.
Jialiang Hou and Fei Gao are also with the Differential Robotics Technology Company, Hangzhou 311100, China. This work was supported by the National Key R$\&$D Program of China under Grant No. 2023YFB4706600,
the Zhejiang Provincial Science and Technology Plan Project under Grant No. 2024C01170 and the National Natural Science Foundation of China under Grant No. 62322314.}
\thanks{E-mail: \{jxlin, jlhou25, fgaoa\}@zju.edu.cn}
}

\makeatletter
\newcommand{\teaserfigure}{%
\par\noindent
\begingroup
\centering
\includegraphics[width=\textwidth]{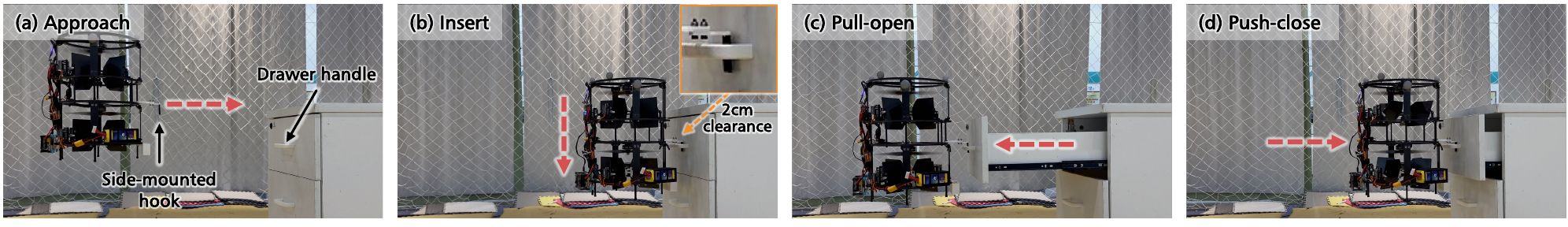}\par
\setlength{\abovecaptionskip}{0pt}%
\setlength{\belowcaptionskip}{0pt}%
\refstepcounter{figure}%
\label{fig:drawer_manipulation}%
\def\@captype{figure}%
\@makecaption{\fnum@figure}{Close-proximity drawer push--pull manipulation with FLOAT Drone.
(a) Approach and alignment using the side-mounted 3-D printed hook.
(b) Hook insertion through the $2~\mathrm{cm}$ handle clearance, with the inset showing the hook--handle interface.
(c) Pull-open motion.
(d) Push-close motion.}
\par
\endgroup
}
\makeatother

\begin{document}

\IEEEaftertitletext{\teaserfigure}
\maketitle


\begin{abstract}
Aerial physical interaction represents a promising direction for next-generation unmanned aerial vehicles (UAVs),
but it requires an aerial platform that can exert contact forces while maintaining stable flight.
For close-proximity tasks, this translates into three coupled design requirements: multidimensional wrench generation for stable contact,
compactness for maneuverability and safety in confined spaces, and reduced lateral airflow toward the target when generating horizontal force.
This article presents FLOAT Drone, a fully actuated coaxial UAV with servo-driven control surfaces for close-proximity physical interaction.
The coaxial dual-rotor layout provides a compact propulsion layout, while the control surfaces, immersed in the rotor downwash, generate lateral forces and moments for 6-DoF wrench generation.
A force-matched computational fluid dynamics (CFD) comparison with a tilted-rotor alternative quantifies the reduction in target-facing lateral airflow.
To account for nonlinear rotor--control-surface coupling in the rotor wake, a high-fidelity polynomial aerodynamic wrench model is identified from precision force measurements and embedded in a constrained nonlinear allocator for real-time wrench tracking.
Comparative flight and interaction experiments show that the proposed framework improves control accuracy over linear allocation baselines, rejects ground-effect and payload disturbances, and enables close-proximity drawer push--pull manipulation through a $2~\mathrm{cm}$ handle clearance.

\end{abstract}

\begin{IEEEkeywords}
Aerial physical interaction, fully actuated UAV, adaptive control.
\end{IEEEkeywords}

\section{Introduction}
\IEEEPARstart{A}{erial} physical interaction is an important direction for next-generation UAVs,
enabling aerial robots to move beyond passive sensing and inspection \cite{zhou2023racer,aucone2023dna} toward active manipulation, transportation, and maintenance tasks \cite{wu2026hand,jin2025tethered}.
Unlike free-flight missions, physical interaction requires a UAV to maintain stable flight while exerting contact forces on the environment,
which places stringent requirements on both the platform design and the control system.
Stable contact requires the platform to generate and regulate multidimensional wrenches,
while close-proximity operation in confined spaces further demands a compact size for maneuverability and safety.
In addition, when horizontal forces are produced by tilting the vehicle or redirecting rotor thrust,
part of the rotor-induced airflow can be directed toward the target, causing undesirable aerodynamic interference.
Therefore, an interaction-oriented aerial platform should combine decoupled wrench generation, compactness, and reduced lateral airflow when generating horizontal forces.

Conventional multirotor UAVs are inherently underactuated because their net thrust is primarily generated along the body-fixed vertical axis \cite{wu2026SR}. As a result, producing horizontal forces requires tilting the vehicle, which couples force regulation with attitude motion and makes it difficult to maintain the desired contact force and vehicle attitude simultaneously during physical interaction.
Fully actuated aerial platforms alleviate this limitation by generating 6-DoF wrenches through fixed nonparallel rotor axes or active thrust-vectoring mechanisms \cite{FlyingHand,bodie2021fulltro}. However, these approaches often require distributed rotor layouts, additional actuators, or tilting mechanisms, increasing the size, mass, and mechanical complexity of the platform. In addition, thrust-vectoring-based lateral force generation may direct part of the rotor-induced airflow toward the interaction target, causing aerodynamic interference in close-proximity tasks. These limitations motivate a compact fully actuated UAV architecture that retains the ability to generate 6-DoF wrenches while producing lateral force with less airflow directed toward the target.

To address these limitations, our preliminary work \cite{FLOATDrone2025} introduced FLOAT Drone, a fully actuated coaxial UAV that combines a compact coaxial dual-rotor layout with servo-driven control surfaces immersed in the rotor downwash. This configuration enables 6-DoF wrench generation by using the control surfaces to generate lateral forces and moments, rather than relying on vehicle tilting or rotor thrust vectoring. It is therefore expected to reduce the target-facing lateral airflow during horizontal force generation. However, the preliminary study left several key issues unresolved: the reduction in lateral airflow was not quantitatively benchmarked against thrust-vectoring baselines, the simplified linear aerodynamic model could not capture the nonlinear rotor-control-surface coupling, and the controller did not explicitly compensate for model uncertainties and contact-induced disturbances.

Building on the preliminary design, this article addresses these unresolved issues through the following contributions:
\begin{enumerate}
    \item We establish a force-matched CFD benchmark to quantify the reduction in target-facing lateral airflow achieved by the rotor--control-surface configuration relative to a tilted-rotor alternative.

    \item We characterize the nonlinear rotor--control-surface aerodynamic coupling and identify a high-fidelity polynomial aerodynamic wrench model from high-precision force measurements over the actuation envelope.

    \item We develop a saturation-aware nonlinear optimization-based control allocator that embeds the identified aerodynamic model for online 6-DoF wrench tracking under actuator constraints.

    \item We augment SE(3)-based geometric control with $\mathcal{L}_1$ adaptive compensation to reject residual model errors, payload variations, ground-effect disturbances, and external disturbances during flight.

    \item We validate the upgraded prototype at the system level, showing that the combined mechatronic design, aerodynamic model, nonlinear allocator, and adaptive controller improve flight accuracy and disturbance robustness for close-proximity physical interaction.
\end{enumerate}

\section{Related Work}

\subsection{Fully Actuated UAV Configurations}
Fully actuated multirotor UAVs are commonly designed by arranging rotor thrust directions in nonparallel configurations. According to whether the rotor axes are fixed or actively changed during flight, existing platforms can be broadly divided into fixed-tilt and variable-tilt configurations~\cite{FullyActuatedReview}.

Fixed-tilt configurations achieve 6-DoF wrench generation by presetting the installation angles of the rotors.
For instance, Rajappa et al.~\cite{HexaRajappa} modified a standard hexacopter by rigidly mounting six rotors with tilted axes in a single plane.
Brescianini and D'Andrea~\cite{OctaDario} achieved omnidirectional flight by placing eight rotors at the vertices of a cubic structure.
Similarly, Park et al.~\cite{OctaPark} proposed a bar-shaped octarotor platform for aerial manipulation while mitigating inter-rotor aerodynamic interference.
Although mechanically simple, fixed-tilt configurations generally suffer from reduced hovering efficiency because only part of the generated thrust contributes to weight compensation, while the lateral thrust components must cancel each other during hover.
Moreover, fixed-tilt platforms often require larger rotor spacing to alleviate inter-rotor wake interference caused by the tilted rotor flows, leading to bulky size and limited suitability for confined-space operation.

Variable-tilt configurations actively adjust rotor orientations during flight, offering a larger feasible wrench set and improved hovering efficiency.
Ryll et al.~\cite{QuadRyll} augmented a standard quadrotor framework by equipping each motor with an independent tilting servo mechanism.
Zheng et al.~\cite{TiltDrone} developed a dual-servo-driven gimbal mechanism to synchronously change the orientation of four rotors.
Kamel et al.~\cite{Voliro} employed six tilting units on a hexacopter platform, while Allenspach et al.~\cite{Twelveallenspach} proposed a coaxial 12-rotor omnidirectional vehicle to improve thrust capability and cancel reaction torques.
Guan et al.~\cite{OctaGuan} investigated a coaxial octarotor combined with four tilting servos.
Compared with fixed-tilt designs, variable-tilt platforms can improve efficiency, but the additional actuators and tilting mechanisms increase mass, size, and mechanical complexity.

Fixed- and variable-tilt platforms generate lateral forces mainly through lateral components of rotor thrust, which may direct rotor-induced flow toward nearby targets in close-proximity tasks. FLOAT Drone instead uses servo-driven control surfaces in the rotor downwash to produce lateral forces within a compact coaxial layout. The resulting rotor--control-surface coupling motivates the high-fidelity modeling and nonlinear allocation developed in this article.
\subsection{Aerial Physical Interaction Control}
Stable aerial physical interaction requires a UAV to regulate its motion while compensating for contact forces, model uncertainties, and aerodynamic disturbances. SE(3) geometric control~\cite{GeoControlLee,MinimumSnapMellinger} provides a coordinate-free formulation suitable for large-attitude maneuvers and singularity-free trajectory tracking. However, in its nominal form, it does not explicitly estimate or compensate for external disturbances and unmodeled aerodynamic effects. Impedance and force control methods~\cite{ImpedanceBodie,ImageHe} are widely used for compliant aerial interaction, but many implementations require force sensing or contact-force estimation, increasing payload and system complexity for compact aerial platforms.

To compensate for disturbances without additional force/torque sensing, disturbance-observer-based (DOB) methods~\cite{DOBReview,DOBquadXuchao,NDOBChen} have been widely studied. These methods can estimate external disturbances from motion and input measurements, but their performance is affected by the trade-off between estimation bandwidth and noise sensitivity.
Incremental Nonlinear Dynamic Inversion (INDI)~\cite{INDITal,INDISun} also provides strong disturbance rejection, but it typically requires reliable acceleration estimates and actuator-state feedback, such as rotor speeds or servo deflections, which can be challenging for compact platforms with noisy measurements and nonlinear actuator coupling.
Recent data-driven methods~\cite{DiffSimPan} improve adaptation to unknown disturbances, but often require high-quality simulation models or substantial offline training.

$\mathcal{L}_1$ adaptive control~\cite{L1QuadWu,FlyingHand} offers an attractive alternative by separating fast adaptation from robust control through a low-pass-filtered compensation channel. This allows rapid disturbance estimation while preventing high-frequency estimation noise from being directly injected into the control input. In this article, we combine SE(3)-based geometric control with $\mathcal{L}_1$ adaptive compensation to construct a robust control framework for FLOAT Drone that compensates for residual model errors and external disturbances without additional force/torque sensors.
\section{Mechatronic Design}
\label{sec:design}

This section presents the mechatronic design of FLOAT Drone, including the compact coaxial rotor configuration, the control-surface-based lateral force generation mechanism, the force-matched CFD evaluation of target-facing lateral airflow, the nominal 6-DoF wrench generation principle, and the prototype implementation.

\begin{figure}[t]
	\centering
	\includegraphics[width=0.75\linewidth]{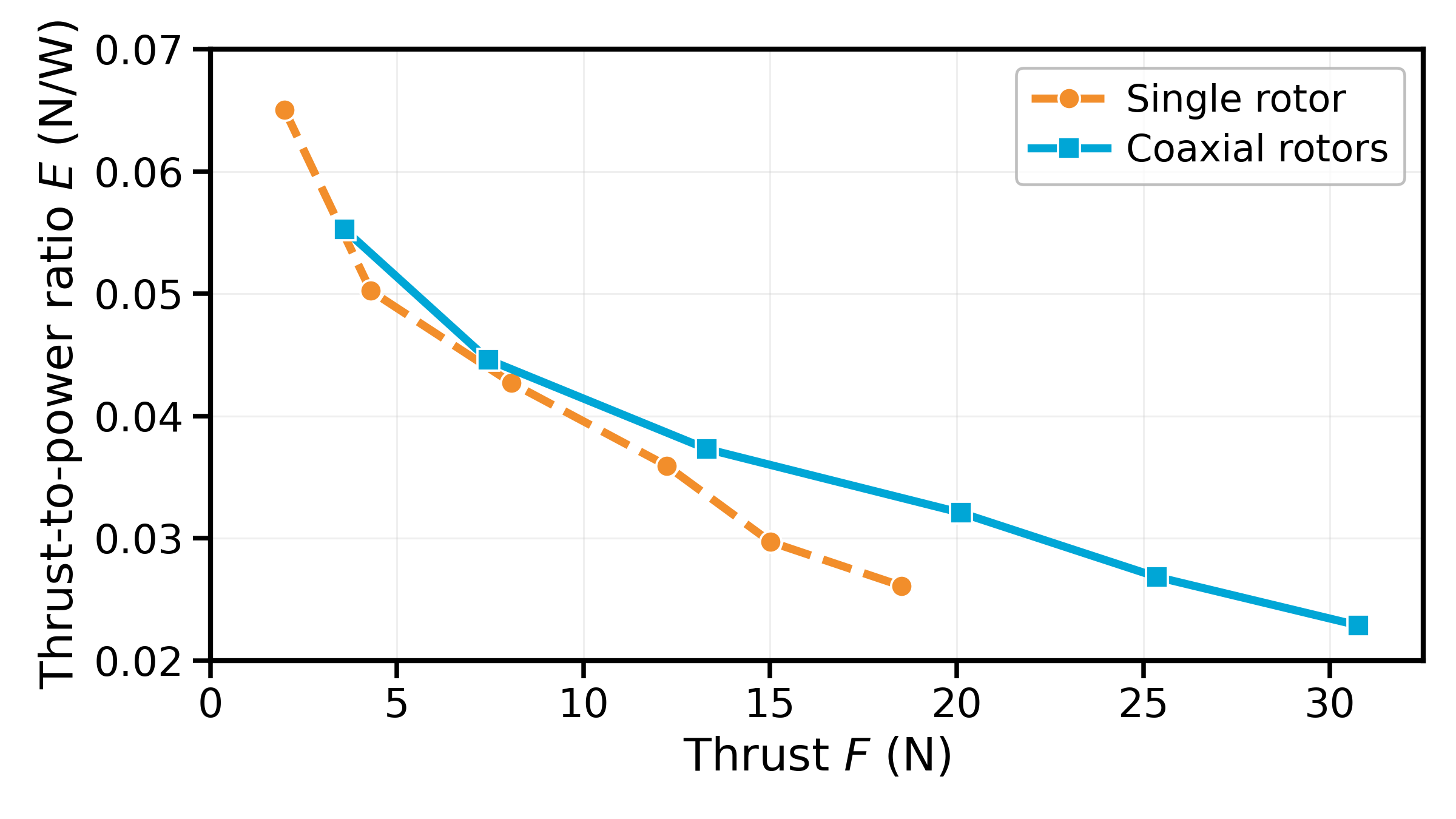}
	\caption{Thrust-to-power comparison of the single-rotor and coaxial dual-rotor configurations.}
	\label{fig:coaxial_efficiency}
\end{figure}

\subsection{Compact and Efficient Coaxial Rotor Configuration}

For close-proximity physical interaction, the horizontal size of the UAV directly affects its maneuverability and accessibility in confined spaces. Following the compactness analysis in our preliminary work~\cite{FLOATDrone2025}, the effective size of a rotor layout is evaluated by the minimum circumscribed area of its horizontal projection. Under the same vehicle weight and hovering-efficiency requirement, concentrating the propulsion system on a single rotor axis yields the smallest horizontal size among conventional multirotor layouts. This motivates the use of a single-axis propulsion architecture as the basis of FLOAT Drone.

Although a single-axis layout is compact, using only one rotor would leave limited thrust margin and payload capacity. FLOAT Drone therefore adopts a coaxial dual-rotor configuration, in which two rotors are vertically stacked along the same axis to improve propulsion performance without significantly increasing the horizontal size. Since the two rotors share nearly the same horizontal projection, the coaxial arrangement preserves the compactness of the single-axis layout while increasing the available thrust.

To evaluate this design choice, we experimentally measured the thrust-to-power characteristics of a single-rotor unit and a coaxial dual-rotor unit. In both tests, the throttle command was swept from 20\% to 70\%. The thrust-to-power ratio is defined as
\begin{equation}
    E = \frac{T}{P_e},
\end{equation}
where $T$ is the generated thrust and $P_e$ is the power consumption.

As shown in Fig.~\ref{fig:coaxial_efficiency}, the coaxial dual-rotor configuration achieves higher propulsion efficiency at the target operating point. At the prototype hovering thrust of approximately $18.1~\mathrm{N}$, its thrust-to-power ratio is 24\% higher than that of the single-rotor configuration. Moreover, under the same 70\% throttle command, the coaxial dual-rotor unit generates 66\% more maximum thrust, providing a larger payload margin. Therefore, the coaxial dual-rotor layout offers a favorable combination of compactness, efficiency, and thrust capability for FLOAT Drone.

\begin{figure}[t]
	\centering
	\includegraphics[width=1.0\linewidth]{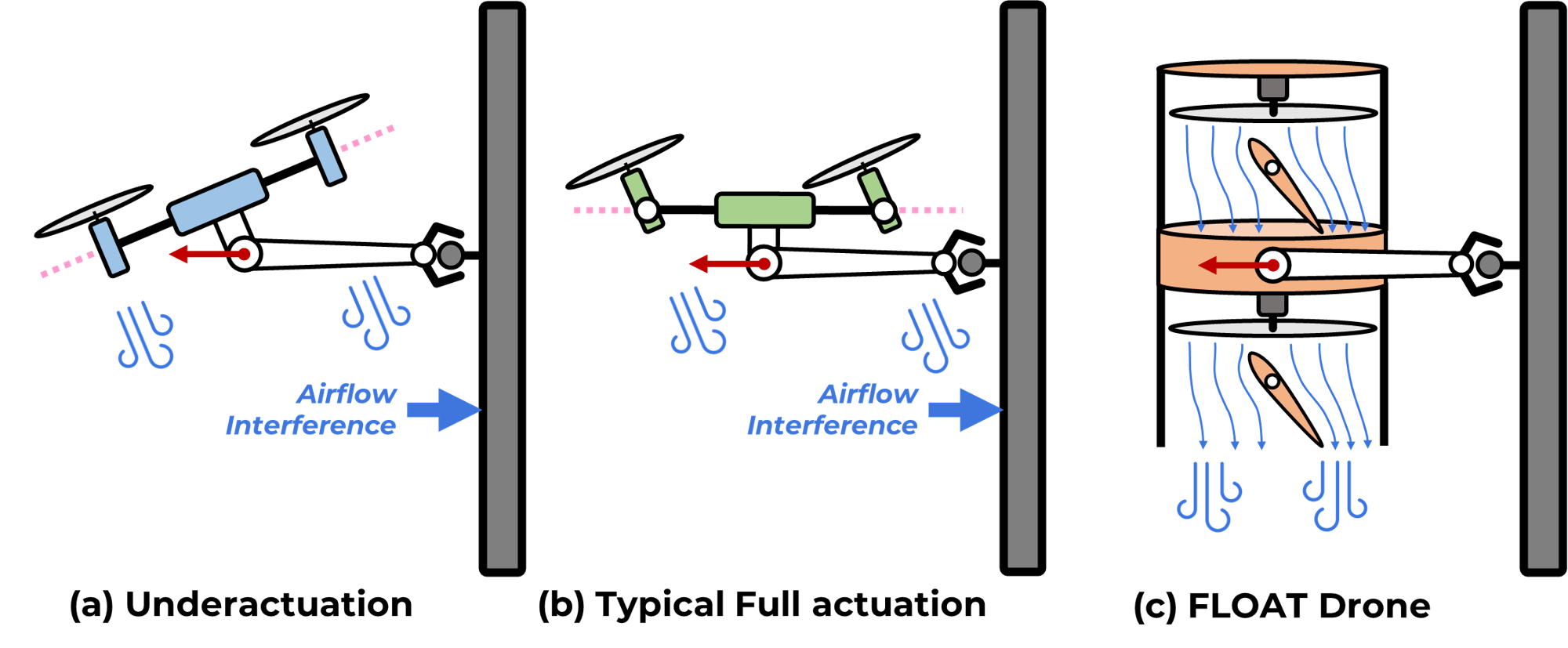}
\caption{Comparison of lateral-force generation mechanisms.
(a) Body tilting in an underactuated UAV.
(b) Rotor-thrust redirection in a tilted-rotor fully actuated UAV.
(c) Control-surface-based lateral-force generation in the rotor wake of FLOAT Drone.}
	\label{fig:lateral_force_generation}
\end{figure}

\subsection{Control-Surface-Based Lateral Force Generation}

The compact coaxial rotor configuration provides vertical thrust within a small horizontal size, but independent lateral force generation requires an additional mechanism. Fig.~\ref{fig:lateral_force_generation} compares three representative approaches: body tilting in underactuated UAVs, rotor-thrust redirection in tilted-rotor fully actuated UAVs, and the control-surface-based mechanism adopted by FLOAT Drone. Instead of tilting the vehicle body or the rotor axes, FLOAT Drone places servo-driven control surfaces in the rotor wake. When a control surface is deflected, the high-speed downwash changes the local pressure distribution around the surface and produces an aerodynamic reaction force. Its lateral component generates horizontal force, while its vertical component introduces thrust loss and aerodynamic coupling.

Since the rotor axes remain aligned with the body vertical axis, the main wake is kept predominantly downward, and lateral force is generated through local wake deflection around the control surfaces. Compared with tilted-rotor thrust vectoring, this mechanism is expected to reduce, but not eliminate, target-facing lateral airflow. The following subsection quantifies this effect through force-matched CFD simulations.

\subsection{Force-Matched CFD Evaluation of Target-Facing Lateral Airflow}
\label{sec:force_matched_cfd}

To quantify the lateral airflow induced during horizontal force generation, a force-matched CFD comparison was conducted between a rotor--control-surface unit and a tilted-rotor baseline. Both units use the same 8-inch three-blade propeller. For the rotor--control-surface unit, a $200~\mathrm{mm}\times100~\mathrm{mm}\times2~\mathrm{mm}$ control surface is placed in the rotor downwash, with its center located on the rotor axis and $100~\mathrm{mm}$ below the rotor center. A virtual target-facing measurement plane is placed parallel to the global vertical axis and $120~\mathrm{mm}$ away from the rotor axis for both configurations.

\begin{figure}[t]
	\centering
	\includegraphics[width=1.0\linewidth]{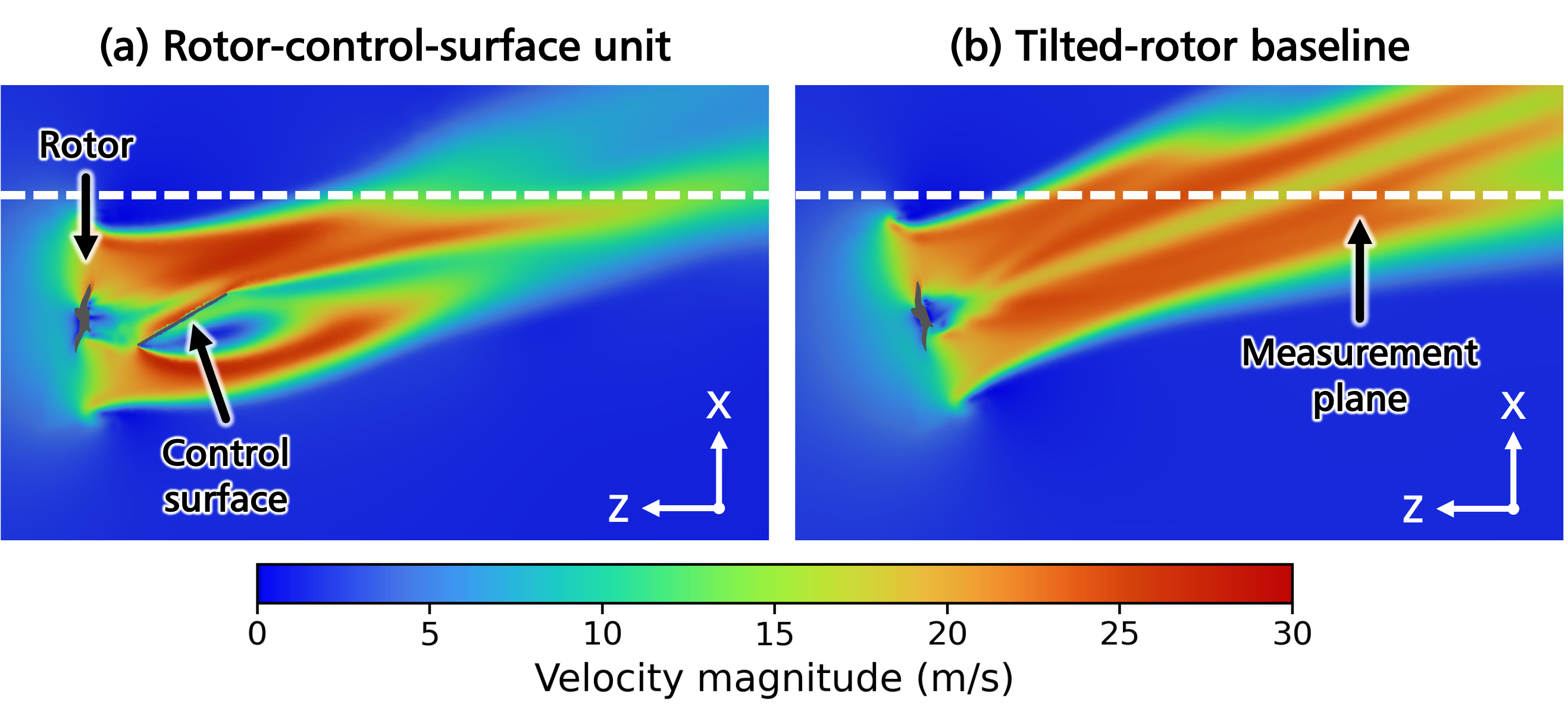}
	\caption{Velocity-magnitude contours from the force-matched CFD comparison at $F_x/F_z=0.35$.
	(a) Rotor--control-surface unit.
	(b) Tilted-rotor baseline.
	The same color scale and view are used in both panels, and the dashed line denotes the target-facing measurement plane $\mathcal A_f$.}
	\label{fig:cfd_setup_flowfield}
\end{figure}

In the rotor--control-surface unit, the rotor speed is fixed at $10000~\mathrm{rpm}$, and the control-surface deflection angle $\delta$ is varied from $0^\circ$ to $45^\circ$ with a $5^\circ$ interval. For each case, the tilted-rotor baseline is tuned by adjusting its rotor speed and inclination angle such that its generated vertical and lateral forces match those of the rotor--control-surface unit. The maximum force-matching error is below $0.5\%$. Fig.~\ref{fig:cfd_setup_flowfield} shows representative velocity contours at $F_x/F_z=0.35$. Under matched force generation, the tilted-rotor baseline redirects the main rotor-induced flow together with the thrust vector, whereas the rotor--control-surface unit keeps the primary wake more vertically oriented and produces lateral force through local wake deflection around the control surface.

\begin{figure}[t]
	\centering
	\includegraphics[width=1.0\linewidth]{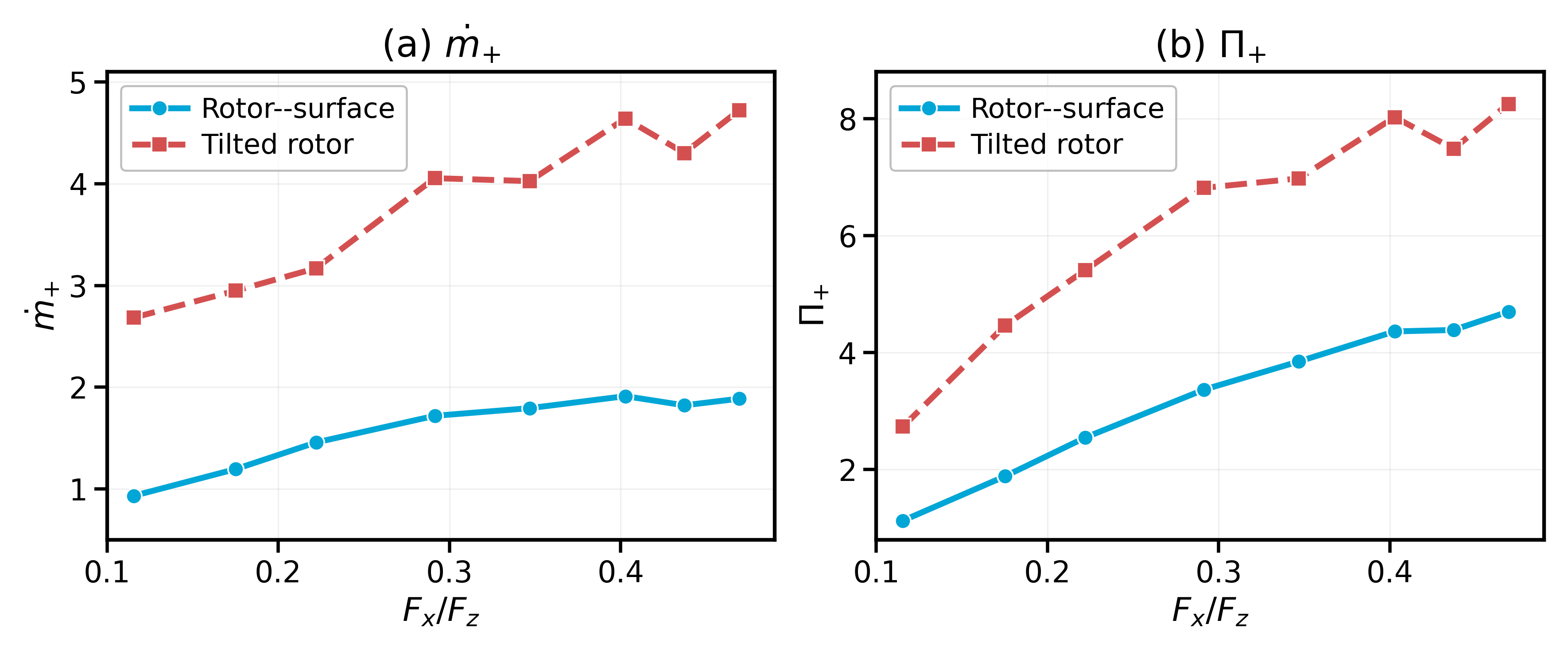}
	\caption{Force-matched CFD comparison of target-facing airflow metrics on $\mathcal A_f$.
	(a) Target-facing mass-flow rate $\dot m_+$.
	(b) Positive-normal momentum-flux integral $\Pi_+$.}
	\label{fig:cfd_airflow_comparison}
\end{figure}

Let $\mathcal{A}_f$ denote the target-facing measurement plane, $\mathbf{n}_f$ its unit normal direction pointing from the vehicle toward the target, $v_n=\mathbf{v}\cdot\mathbf{n}_f$ the normal velocity, and $v_n^+=\max(v_n,0)$ the target-facing normal component. The target-facing mass-flow rate and the corresponding normal momentum-flux integral are computed as
\begin{equation}
    \dot{m}_+=\int_{\mathcal{A}_f}\rho v_n^+ dA,
    \qquad
    \Pi_+=\int_{\mathcal{A}_f}\rho (v_n^+)^2 dA .
\end{equation}
Here, $\dot{m}_+$ accounts only for lateral airflow directed toward the target, while $\Pi_+$ measures the associated positive-normal momentum flux across the virtual plane and should not be interpreted as the aerodynamic force acting on a physical target.

The quantitative results are shown in Fig.~\ref{fig:cfd_airflow_comparison}. The cases with $\delta=0^\circ$ and $5^\circ$ serve as cases with nearly zero lateral force and are omitted from the summary plot. For the nonzero lateral-force cases from $\delta=10^\circ$ to $45^\circ$, the rotor--control-surface unit consistently yields lower $\dot{m}_+$ and $\Pi_+$ than the tilted-rotor baseline. On average, the target-facing mass-flow rate is reduced by $58.5\%$, and the target-facing normal momentum-flux integral is reduced by $49.4\%$. These results indicate that, under matched vertical and lateral force generation, the rotor--control-surface mechanism reduces target-facing lateral airflow compared with tilted-rotor thrust vectoring.
\begin{figure}[t]
	\centering
	\includegraphics[width=1.0\linewidth]{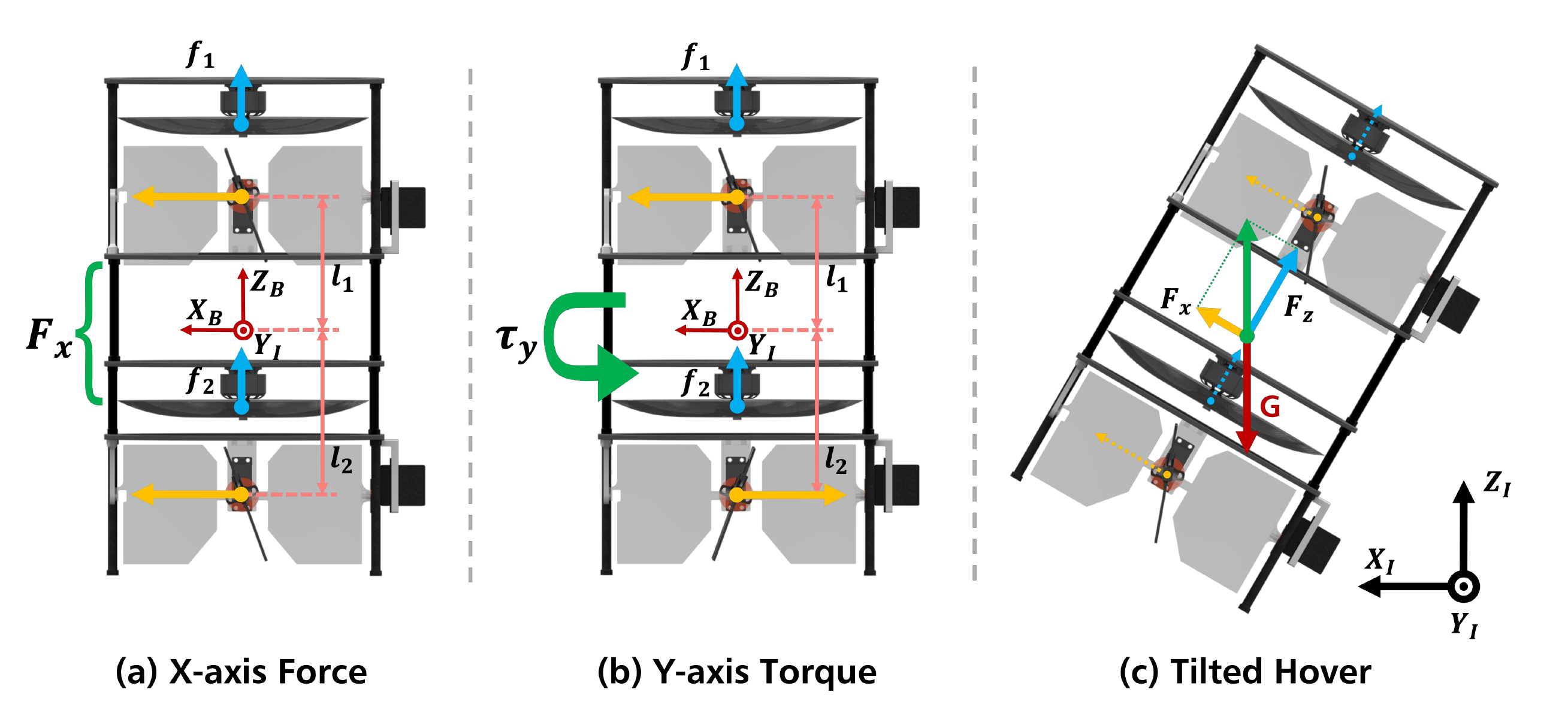}
	\caption{Nominal 6-DoF wrench-generation principle of FLOAT Drone. Common lateral forces generate net horizontal force, whereas differential upper/lower lateral forces generate roll/pitch torque through vertical separation.}
	\label{fig:wrench_generation}
\end{figure}
\subsection{Nominal 6-DoF Wrench Generation Mechanism}

The compact coaxial rotor layout and the control-surface-based lateral force generation mechanism together provide a nominal basis for 6-DoF wrench generation, as illustrated in Fig.~\ref{fig:wrench_generation}. FLOAT Drone consists of two counter-rotating coaxial rotors and four servo-driven control surfaces arranged in two vertically separated rotor-control-surface units. The coaxial rotors mainly generate vertical force and yaw torque: a common change in rotor speeds adjusts the total lift, whereas a differential change adjusts the net reaction torque around the body $z$ axis.

The control surfaces generate the remaining horizontal forces and roll/pitch torques. For each horizontal direction, the upper and lower control surfaces can produce lateral aerodynamic forces at different heights with respect to the center of mass. When the upper and lower surfaces generate lateral forces in the same direction, their moment contributions approximately cancel and a net lateral force is produced. When they generate lateral forces in opposite directions, the force contributions approximately cancel and a roll or pitch torque is produced through the vertical moment arm. Therefore, common and differential deflections of the vertically separated control surfaces provide nominal control of the horizontal forces and roll/pitch torques, respectively.

In summary, the coaxial rotors provide nominal control of $F_z$ and $\tau_z$, while the control surfaces provide nominal control of $F_x$, $F_y$, $\tau_x$, and $\tau_y$. This arrangement enables FLOAT Drone to generate 6-DoF wrenches without relying on large body tilting or rotor-axis tilting. It should be noted that this is only a nominal mechanism: in practice, the generated wrench is affected by thrust loss, rotor wake distortion, and aerodynamic coupling between the rotors and control surfaces. These nonlinear effects are modeled and compensated in the following sections.

\subsection{High-Payload Prototype Implementation}
\label{sec:prototype}

\begin{figure}[t]
	\centering
	\includegraphics[width=0.9\linewidth]{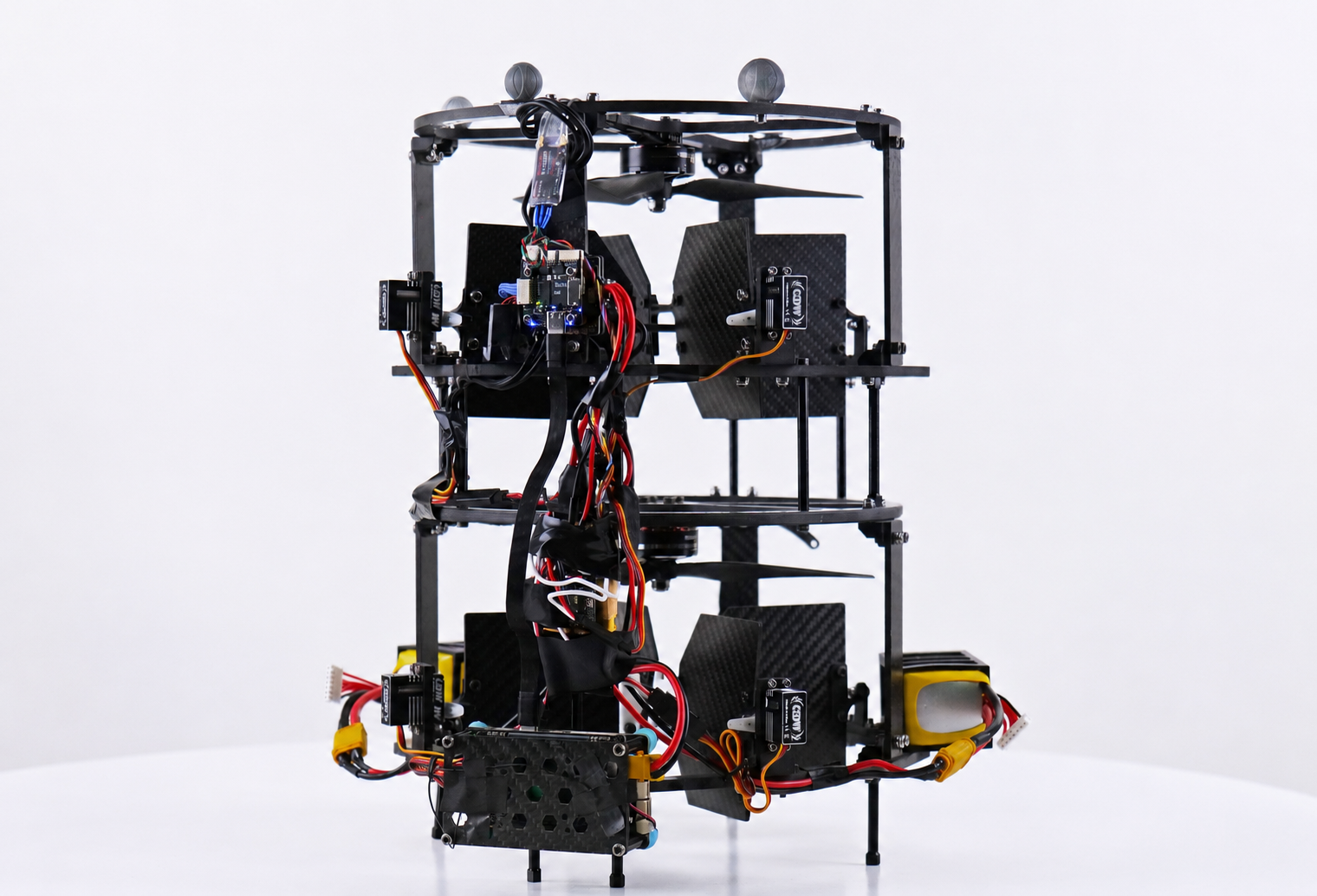}
	\caption{Upgraded FLOAT Drone prototype.}
	\label{fig:hardware}
\end{figure}

Compared with the preliminary prototype~\cite{FLOATDrone2025}, this work implements an upgraded FLOAT Drone platform for aerial manipulation tasks. The purpose of this redesign is to provide additional payload margin for end-effectors, such as lightweight manipulators or grippers, while preserving the coaxial rotor-control-surface architecture described above. The prototype is shown in Fig.~\ref{fig:hardware}, and its main hardware components are summarized in Table~\ref{tab:prototype_specs}.

\begin{table}[t]
\centering
\caption{Main Hardware Components of the Upgraded Prototype.}
\label{tab:prototype_specs}
\small
\renewcommand{\arraystretch}{1.05}
\begin{tabular*}{\linewidth}{@{\extracolsep{\fill}}ll@{}}
\toprule
Component & Model \\
\midrule
Propeller & HQProp 8045-3 8-inch \\
Motor & T-MOTOR V3008 1350KV \\
ESC & HAKRC 65A \\
Control-surface servo & GDW DS290MG digital servo \\
Battery & GNB 6S 1300-mAh 160C LiPo \\
Flight controller & NxtPX4v2 \\
Onboard computer & NVIDIA Jetson Orin NX \\
\bottomrule
\end{tabular*}
\end{table}

The final prototype has a total mass of approximately $1.85~\mathrm{kg}$ and an overall size of $265~\mathrm{mm}\times310~\mathrm{mm}\times340~\mathrm{mm}$, with a horizontal circumscribed diameter of approximately $312~\mathrm{mm}$. The maximum continuous hover duration measured in flight tests is $225~\mathrm{s}$. All flight and interaction experiments in the following sections are conducted using this upgraded prototype.

\section{High-Fidelity Aerodynamic Modeling and Control Allocation}
\label{sec:model_allocation}

The nominal wrench-generation mechanism shown in Fig.~\ref{fig:wrench_generation} explains how FLOAT Drone can produce 6-DoF wrenches in principle. However, because the control surfaces operate inside the coaxial rotor wake, the actual actuator-to-wrench mapping contains thrust loss, nonlinear surface forces, and cross-coupling among rotors and control surfaces. This section first introduces the nominal rigid-body dynamics and the linear allocation baseline, then analyzes the aerodynamic coupling using CFD, identifies a high-fidelity aerodynamic wrench model from force measurements, and finally embeds the identified model into a nonlinear optimization-based control allocator.

\subsection{Nominal Dynamics and Linear Allocation Baseline}

Let $\mathcal F_I$ and $\mathcal F_B$ denote the inertial frame and the body-fixed frame, respectively. The rigid-body dynamics of FLOAT Drone are
\begin{equation}
    m\ddot{\mathbf p}=m\mathbf g+\mathbf R\mathbf T_B,
    \qquad
    \mathbf J\dot{\boldsymbol\omega}
    =
    -\boldsymbol\omega\times\mathbf J\boldsymbol\omega+\boldsymbol\tau_B ,
\end{equation}
where $\mathbf g=[0,0,-g]^T$ is the gravity acceleration vector expressed in $\mathcal F_I$, and $\mathbf T_B$ and $\boldsymbol\tau_B$ are the force and torque generated by the propulsion and control-surface system in the body frame. We define the body-frame aerodynamic wrench as
\begin{equation}
    \mathbf w =
    [\mathbf T_B^T,\boldsymbol\tau_B^T]^T .
\end{equation}

The linear allocation baseline follows a simplified model derived from ideal rotor momentum relations and small-angle control-surface lift approximations. This model assumes that rotor thrust controls $F_z$ and $\tau_z$, while the control-surface lift is linearly proportional to the surface deflection and local downwash velocity. Under these assumptions, a desired wrench command can be mapped to actuator commands through a closed-form linear allocator. The explicit closed-form allocation equations are given in our preliminary work~\cite{FLOATDrone2025}. In practice, however, this model neglects thrust loss, wake distortion, and rotor--control-surface cross-coupling, which motivates the high-fidelity model identified below.

\begin{figure}[t]
	\centering
	\includegraphics[width=1.0\linewidth]{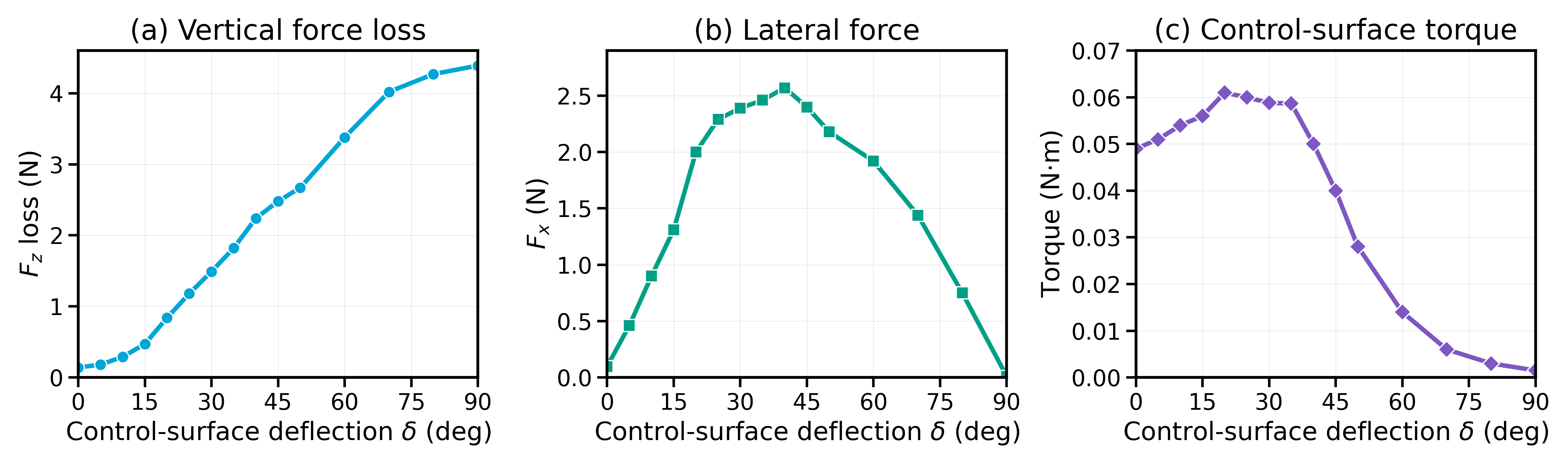}
\caption{CFD analysis of rotor--control-surface coupling. Control-surface deflection produces lateral force, vertical force loss, and control-surface torque.}
\label{fig:cfd_coupling}
\end{figure}

\subsection{CFD Analysis of Rotor--Control-Surface Coupling}
\label{sec:cfd_coupling}

To illustrate the aerodynamic coupling introduced by the control surfaces, a representative single-rotor--single-control-surface unit is analyzed using CFD. The setup is similar to that in the force-matched CFD study in Section~\ref{sec:force_matched_cfd}. In this analysis, the rotor speed is fixed at $10000~\mathrm{rpm}$, producing approximately $10~\mathrm{N}$ of thrust and $-0.15~\mathrm{N\,m}$ of rotor torque, while the control-surface deflection angle is swept from $0^\circ$ to $90^\circ$. The focus here is on the force and moment generated on the control surface itself.

As shown in Fig.~\ref{fig:cfd_coupling}, the lateral force generated by the control surface varies nonlinearly with the deflection angle. It increases at small and moderate deflections, reaches a maximum around $40^\circ$, and then decreases as the deflection further increases. Meanwhile, the control surface produces a vertical force component opposite to the rotor thrust direction, which appears as thrust loss and increases with the deflection angle. The associated control-surface torque also varies with the deflection angle. These results show that control-surface-based lateral force generation is coupled with vertical thrust loss and control-surface torque variation, motivating the high-fidelity aerodynamic wrench model identified in the next subsection.

\begin{figure}[t]
	\centering
	\includegraphics[width=1.0\linewidth]{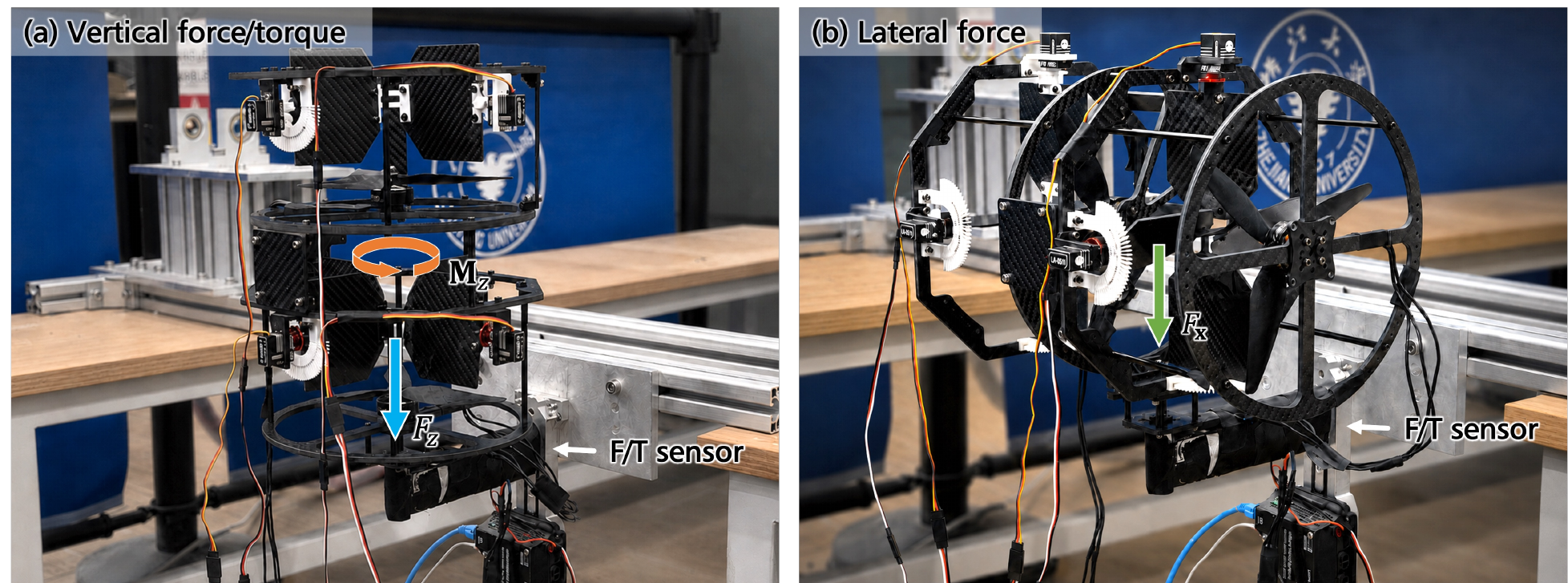}
	\caption{Static force-measurement setups for aerodynamic model identification.
	(a) Vertical-force and yaw-moment measurement.
	(b) Lateral-force measurement with the test unit reoriented to align the generated force with the sensor axis.}
	\label{fig:force_measure}
\end{figure}

\subsection{High-Fidelity Aerodynamic Wrench Model Identification}
\label{sec:model_identification}

To obtain an actuator-to-wrench model for real-time nonlinear allocation, a high-fidelity aerodynamic model is identified from static force-measurement data, as shown in Fig.~\ref{fig:force_measure}. The motor throttle commands are sampled within $40\%$--$60\%$, and the control-surface deflection angles are sampled within $0^\circ$--$45^\circ$, covering the main operating region of FLOAT Drone.

Let $\mathbf u=[\eta_1,\eta_2,\theta_1,\theta_2,\delta_1,\delta_2]^T$ denote the actuator command vector, where $\eta_1$ and $\eta_2$ are the upper and lower rotor throttle commands, and $\theta_i,\delta_i$ are the control-surface deflections. The aerodynamic wrench is modeled as
\begin{equation}
    \hat{\mathbf w}_{\rm aero}(\mathbf u)
    =
    [\hat F_x,\hat F_y,\hat F_z,\hat \tau_x,\hat \tau_y,\hat \tau_z]^T,
\end{equation}
where each component is represented by a pruned second-order polynomial:
\begin{equation}
    \hat w_i(\mathbf u)
    =
    \sum_{k=1}^{M_i} c_{ik}\phi_k(\mathbf u),
    \qquad i=1,\ldots,6 .
\end{equation}
Here, $\phi_k(\mathbf u)$ denotes a polynomial basis term, and insignificant terms are removed after fitting to reduce the computational cost of online optimization.

For $F_z$ and $\tau_z$, approximately 600 steady-state samples are collected over the above actuation envelope to capture the vertical force loss and control-surface-induced vertical-axis torque caused by control-surface deflection. For horizontal force modeling, only the $x$-axis force is experimentally identified, and the $y$-axis counterpart is obtained from the geometric symmetry of the vehicle. Since the force sensor measures the total horizontal force, a differential measurement scheme is used to separate the upper and lower control-surface contributions. With the lower surface unchanged, reversing the upper surface deflection gives two measurements $F_x^+$ and $F_x^-$. Based on the approximate odd symmetry $F(\delta)\approx -F(-\delta)$, the two contributions are estimated as
\begin{equation}
    F_x^u \approx \frac{F_x^{+}-F_x^{-}}{2},
    \qquad
    F_x^l \approx \frac{F_x^{+}+F_x^{-}}{2}.
\end{equation}

\begin{figure}[t]
	\centering
	\includegraphics[width=0.9\linewidth]{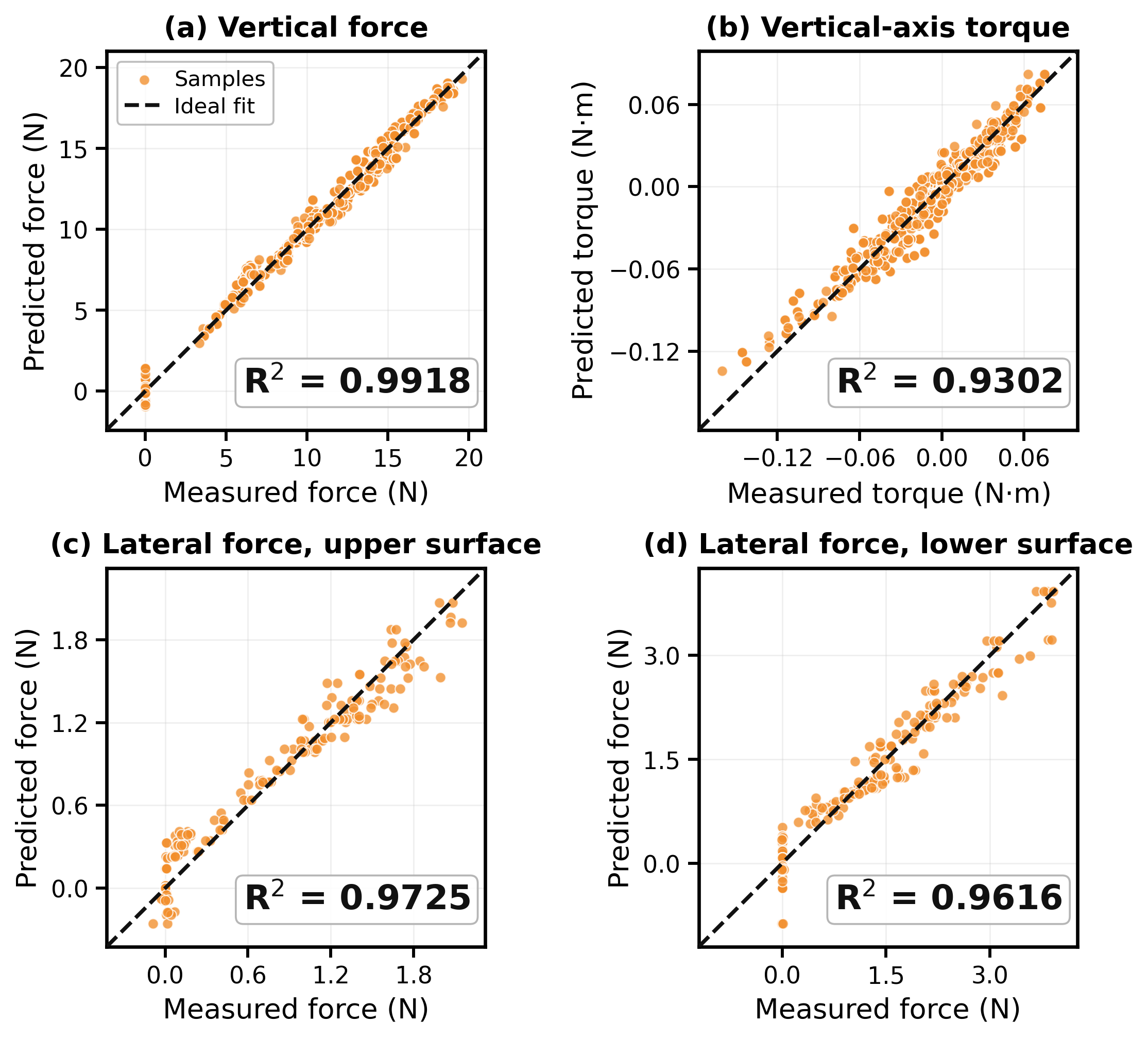}
	\caption{Measured-versus-predicted validation of the identified aerodynamic wrench model.
	(a) Vertical force.
	(b) Vertical-axis torque.
	(c) Upper-surface lateral force.
	(d) Lower-surface lateral force.}
	\label{fig:R2}
\end{figure}

Polynomial models are then fitted for the vertical force, yaw moment, and separated horizontal force components. The corresponding roll and pitch torques are computed from the separated upper/lower lateral forces and their vertical moment arms according to the nominal wrench-generation geometry. The measured-versus-predicted comparisons in Fig.~\ref{fig:R2} show that the identified models achieve $R^2>0.93$ over the tested operating range, indicating that the dominant nonlinear aerodynamic coupling is captured. The complete aerodynamic wrench model is obtained by combining these identified components with the nominal wrench-generation structure in Section~\ref{sec:design}, and is used as the prediction model in the nonlinear control allocator developed next.

\subsection{Model-Based Nonlinear Control Allocation}
\label{sec:nonlinear_allocation}

The identified aerodynamic wrench model provides a nonlinear mapping from actuator commands to the generated 6-DoF wrench. Therefore, the desired wrench from the upper-level controller cannot be accurately allocated using a closed-form linear mixer. Instead, we formulate control allocation as a constrained nonlinear least-squares problem.

Let $\mathbf w_c=[\mathbf T_c^T,\boldsymbol{\tau}_c^T]^T$ denote the desired body-frame wrench command supplied by the upper-level controller, and let $\hat{\mathbf w}_{\rm aero}(\mathbf u)$ be the identified aerodynamic model in Section~\ref{sec:model_identification}. The optimal actuator command is obtained by
\begin{equation}
\begin{aligned}
    \mathbf u^\star
    =
    \arg\min_{\mathbf u}\quad
    &
    \left\|
    \hat{\mathbf w}_{\rm aero}(\mathbf u)-\mathbf w_c
    \right\|_2^2 \\
    \mathrm{s.t.}\quad
    &
    \mathbf u_{\min}\leq \mathbf u \leq \mathbf u_{\max},
\end{aligned}
\label{eq:nlca}
\end{equation}
where $\mathbf u_{\min}$ and $\mathbf u_{\max}$ define the admissible actuator range.

The optimization problem is implemented in CasADi \cite{casadi2019} and solved using IPOPT \cite{ipopt2006implementation}. To ensure real-time execution, the pruned second-order aerodynamic model is used in the allocator, and the solver is generated as C code and integrated into the C++ flight-control program. At each control step, the previous solution is used as the warm-start initial guess.

\begin{figure}[t]
	\centering
	\includegraphics[width=1.0\linewidth]{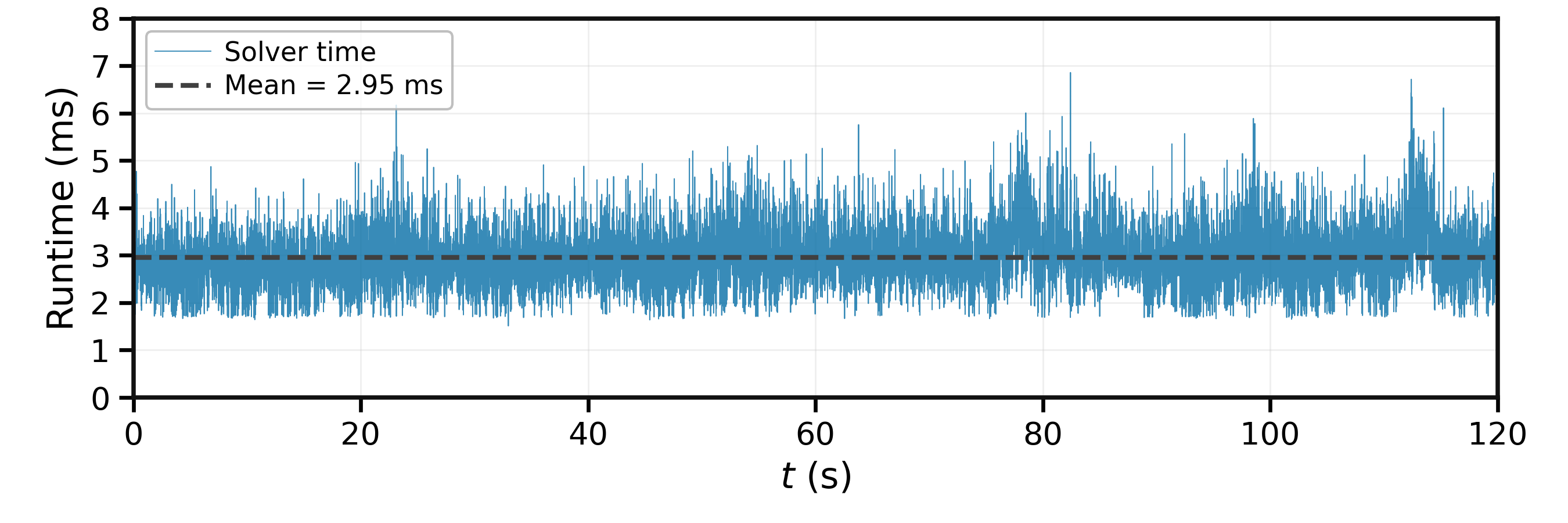}
	\caption{Runtime of the nonlinear control allocator in flight experiments.}
	\label{fig:solver_time}
\end{figure}

The allocator runtime is evaluated during flight experiments with a $100~\mathrm{Hz}$ control loop. As shown in Fig.~\ref{fig:solver_time}, the average solve time is $2.95~\mathrm{ms}$, and the maximum solve time is $6.86~\mathrm{ms}$, demonstrating that the nonlinear allocator met the $100~\mathrm{Hz}$ control frequency requirement in the evaluated flight experiments.

\section{Geometric $\mathcal{L}_1$ Adaptive Control}
\label{sec:l1_control}

\begin{figure}[t]
	\centering
	\includegraphics[width=1.0\linewidth]{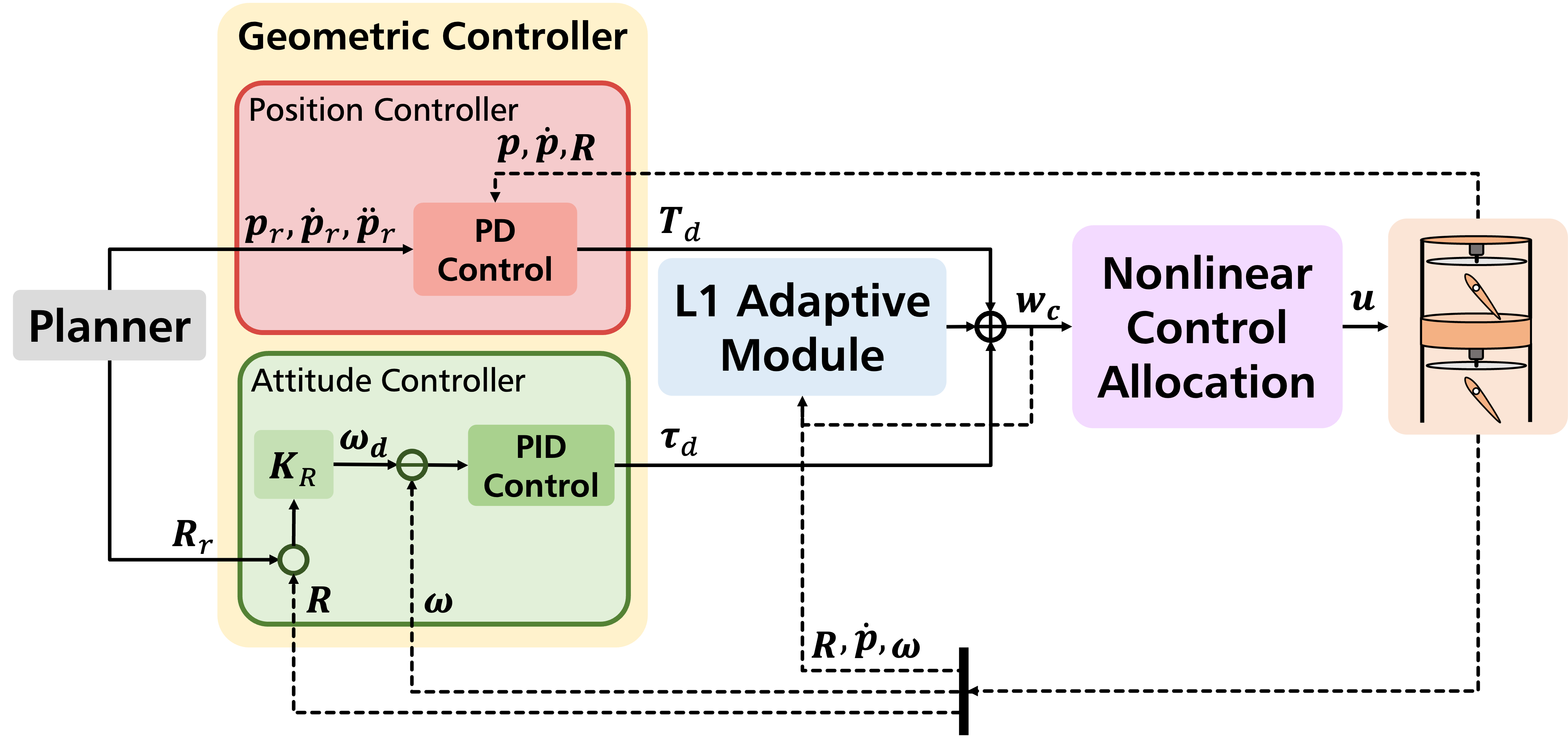}
	\caption{Closed-loop control architecture with geometric control, $\mathcal{L}_1$ adaptive compensation, and nonlinear control allocation.}
	\label{fig:control_structure}
\end{figure}

This section presents the closed-loop control architecture of FLOAT Drone. As shown in Fig.~\ref{fig:control_structure}, the controller consists of an SE(3)-based geometric controller, an $\mathcal{L}_1$ adaptive augmentation module, and the nonlinear control allocator developed in Section~\ref{sec:nonlinear_allocation}. The geometric controller generates a nominal wrench command for trajectory and attitude tracking. The $\mathcal{L}_1$ adaptive module estimates the equivalent lumped disturbance and provides a filtered compensation wrench. The compensated wrench command is then mapped to actuator commands by the nonlinear allocator.

\subsection{Nominal SE(3)-Based Geometric Control}
\label{sec:geometric_control}

The nominal controller consists of a position controller and an attitude controller. Since FLOAT Drone can command three-axis forces and three-axis torques, translational and rotational commands are computed separately at the wrench-command level. Their coupled realization through the rotors and control surfaces is handled by the nonlinear control allocator.

Let $\mathbf p$, $\dot{\mathbf p}$, $\mathbf R$, and $\boldsymbol{\omega}$ denote the position, velocity, attitude, and body angular velocity of the vehicle, respectively. The reference trajectory is given by $\mathbf p_r$, $\dot{\mathbf p}_r$, and $\ddot{\mathbf p}_r$. The desired acceleration is computed as
\begin{equation}
	\ddot{\mathbf p}_d
	=
	\ddot{\mathbf p}_r
	+
	\mathbf K_v
	\left(
	\dot{\mathbf p}_r
	+
	\mathbf K_p
	(\mathbf p_r-\mathbf p)
	-
	\dot{\mathbf p}
	\right),
	\label{eq:desired_acceleration}
\end{equation}
where $\mathbf K_p$ and $\mathbf K_v$ are positive definite gain matrices. The desired body-frame force command is then
\begin{equation}
	\mathbf T_d
	=
	m\mathbf R^T
	\left(
	\ddot{\mathbf p}_d-\mathbf g
	\right),
	\label{eq:desired_force}
\end{equation}
where $m$ is the vehicle mass and $\mathbf g=[0,0,-g]^T$ is the gravity acceleration vector.

The attitude error is defined on SO(3) as
\begin{equation}
	\mathbf R_e
	=
	\frac{1}{2}
	\left(
	\mathbf R_r^T\mathbf R
	-
	\mathbf R^T\mathbf R_r
	\right)^\vee ,
	\label{eq:attitude_error}
\end{equation}
where $\mathbf R_r$ is the desired attitude and $(\cdot)^\vee$ denotes the vee map from $\mathfrak{so}(3)$ to $\mathbb R^3$. The desired angular velocity is generated from the attitude error as $\boldsymbol{\omega}_d=\mathbf K_R\mathbf R_e$, where $\mathbf K_R$ is the attitude gain matrix, and the angular velocity error is defined as $\boldsymbol{\omega}_e=\boldsymbol{\omega}_d-\boldsymbol{\omega}$.
The nominal torque command is generated by a PID controller:
\begin{equation}
	\boldsymbol{\tau}_d
	=
	\mathbf K_{P,\omega}\boldsymbol{\omega}_e
	+
	\mathbf K_{I,\omega}
	\int \boldsymbol{\omega}_e dt
	+
	\mathbf K_{D,\omega}
	\dot{\boldsymbol{\omega}}_e ,
	\label{eq:desired_torque}
\end{equation}
where $\mathbf K_{P,\omega}$, $\mathbf K_{I,\omega}$, and $\mathbf K_{D,\omega}$ are positive definite gain matrices. The nominal wrench command from the geometric controller is therefore

\begin{equation}
    \mathbf w_d =
    [\mathbf T_d^T,\boldsymbol\tau_d^T]^T .
    \label{eq:nominal_wrench}
\end{equation}

\subsection{$\mathcal{L}_1$ Adaptive Disturbance Compensation}
\label{sec:l1_augmentation}

Although the nonlinear allocator improves wrench-tracking accuracy, residual modeling errors and external disturbances still exist during flight and physical interaction, such as ground effect, payload variation, and contact-induced forces. Therefore, an $\mathcal{L}_1$ adaptive module is added to the nominal geometric controller to estimate the equivalent lumped disturbance and generate a filtered compensation wrench.

The input of the $\mathcal{L}_1$ module includes the nominal wrench command $\mathbf w_d$ from the geometric controller, the current attitude $\mathbf R$, and the velocity-level state $\mathbf z=[\dot{\mathbf p}^T,\boldsymbol{\omega}^T]^T$, where $\dot{\mathbf p}$ and $\boldsymbol{\omega}$ are obtained from the onboard state estimator. The derivative $\dot{\mathbf z}$ contains the translational and angular accelerations and is used only in the dynamic model, not as a measured input. The system dynamics are written in the matched-uncertainty form
\begin{equation}
    \dot{\mathbf z}
    =
    \mathbf f(\mathbf z)
    +
    \mathbf B(\mathbf R)
    \left(
        \mathbf w_c+\boldsymbol{\sigma}
    \right),
    \label{eq:l1_actual_dynamics}
\end{equation}
where $\mathbf w_c$ is the compensated wrench command sent to the allocator, and $\boldsymbol{\sigma}$ denotes the equivalent lumped disturbance in the body-frame wrench channel. The known dynamics and input mapping are
\begin{equation}
\begin{aligned}
    \mathbf f(\mathbf z)
    =
    \begin{bmatrix}
        \mathbf g \\
        -\mathbf J^{-1}
        \left(
        \boldsymbol{\omega}\times\mathbf J\boldsymbol{\omega}
        \right)
    \end{bmatrix},
    \\
    \mathbf B(\mathbf R)
    =
    \begin{bmatrix}
        m^{-1}\mathbf R & \mathbf 0_{3\times3} \\
        \mathbf 0_{3\times3} & \mathbf J^{-1}
    \end{bmatrix}.
\end{aligned}
    \label{eq:l1_fb}
\end{equation}
Here, $\mathbf B(\mathbf R)$ maps a body-frame wrench to the corresponding linear and angular accelerations. Thus, external forces, payload changes, residual allocation errors, and unmodeled aerodynamic effects are represented as an equivalent disturbance wrench $\boldsymbol{\sigma}$.

The $\mathcal{L}_1$ module maintains an internal predicted state $\hat{\mathbf z}$, which is not measured but propagated online by the predictor
\begin{equation}
    \dot{\hat{\mathbf z}}
    =
    \mathbf f(\mathbf z)
    +
    \mathbf B(\mathbf R)
    \left(
        \mathbf w_c+\hat{\boldsymbol{\sigma}}
    \right)
    +
    \mathbf A_s
    \left(
        \hat{\mathbf z}-\mathbf z
    \right),
    \label{eq:l1_predictor}
\end{equation}
where $\hat{\boldsymbol{\sigma}}$ is the estimated equivalent disturbance and $\mathbf A_s$ is a Hurwitz matrix. The prediction error is defined as $\tilde{\mathbf z}=\hat{\mathbf z}-\mathbf z$. If the disturbance estimate is accurate, the predicted state should remain close to the measured state; therefore, $\tilde{\mathbf z}$ is used to update $\hat{\boldsymbol{\sigma}}$.

A piecewise-constant adaptation law is used. For $t\in[iT_s,(i+1)T_s)$, the disturbance estimate is held constant as $\hat{\boldsymbol{\sigma}}(t)=\hat{\boldsymbol{\sigma}}(iT_s)$, where
\begin{equation}
    \hat{\boldsymbol{\sigma}}(iT_s)
    =
    -
    \mathbf G(\mathbf R(iT_s))
    \boldsymbol{\Phi}^{-1}
    \boldsymbol{\mu}(iT_s),
    \label{eq:l1_adaptation}
\end{equation}
with $\mathbf G(\mathbf R)=\mathbf B^{-1}(\mathbf R)$,
\begin{equation}
    \boldsymbol{\mu}(iT_s)
    =
    e^{\mathbf A_sT_s}
    \tilde{\mathbf z}(iT_s),
    \qquad
    \boldsymbol{\Phi}
    =
    \mathbf A_s^{-1}
    \left(
        e^{\mathbf A_sT_s}
        -
        \mathbf I
    \right).
    \label{eq:l1_mu_phi}
\end{equation}
This update law converts the predictor error into an equivalent acceleration-level mismatch over one adaptation interval, and then maps it back to the body-frame wrench space through $\mathbf G(\mathbf R)$. In this way, $\hat{\boldsymbol{\sigma}}$ estimates the disturbance wrench that best explains the mismatch between the predicted and measured motion.

The disturbance estimate $\hat{\boldsymbol{\sigma}}$ is not directly used as a control command. Instead, it is passed through a low-pass filter to generate a bandwidth-limited compensation wrench:
\begin{equation}
    \mathbf w_{\mathcal L_1}(s)
    =
    -
    \mathbf C(s)
    \hat{\boldsymbol{\sigma}}(s),
    \label{eq:l1_filter}
\end{equation}
where $\mathbf C(s)$ is a diagonal low-pass filter satisfying $\mathbf C(0)=\mathbf I$. The negative sign indicates that the compensation wrench acts in the opposite direction to the estimated disturbance.

The final wrench command sent to the nonlinear allocator is
\begin{equation}
    \mathbf w_c
    =
    \mathbf w_d
    +
    \mathbf w_{\mathcal L_1}
    =
    \mathbf w_d
    -
    \mathbf C(s)\hat{\boldsymbol{\sigma}}(s).
    \label{eq:l1_compensated_wrench}
\end{equation}
In implementation, the $\mathcal{L}_1$ module reads $\mathbf z$ and $\mathbf R$, compares the internally propagated state $\hat{\mathbf z}$ with the measured state, estimates $\hat{\boldsymbol{\sigma}}$ using \eqref{eq:l1_adaptation}, filters it to obtain $\mathbf w_{\mathcal L_1}$, and sends the compensated wrench $\mathbf w_c$ to the nonlinear control allocator.

\begin{figure}[t]
    \centering
    \includegraphics[width=1.0\linewidth]{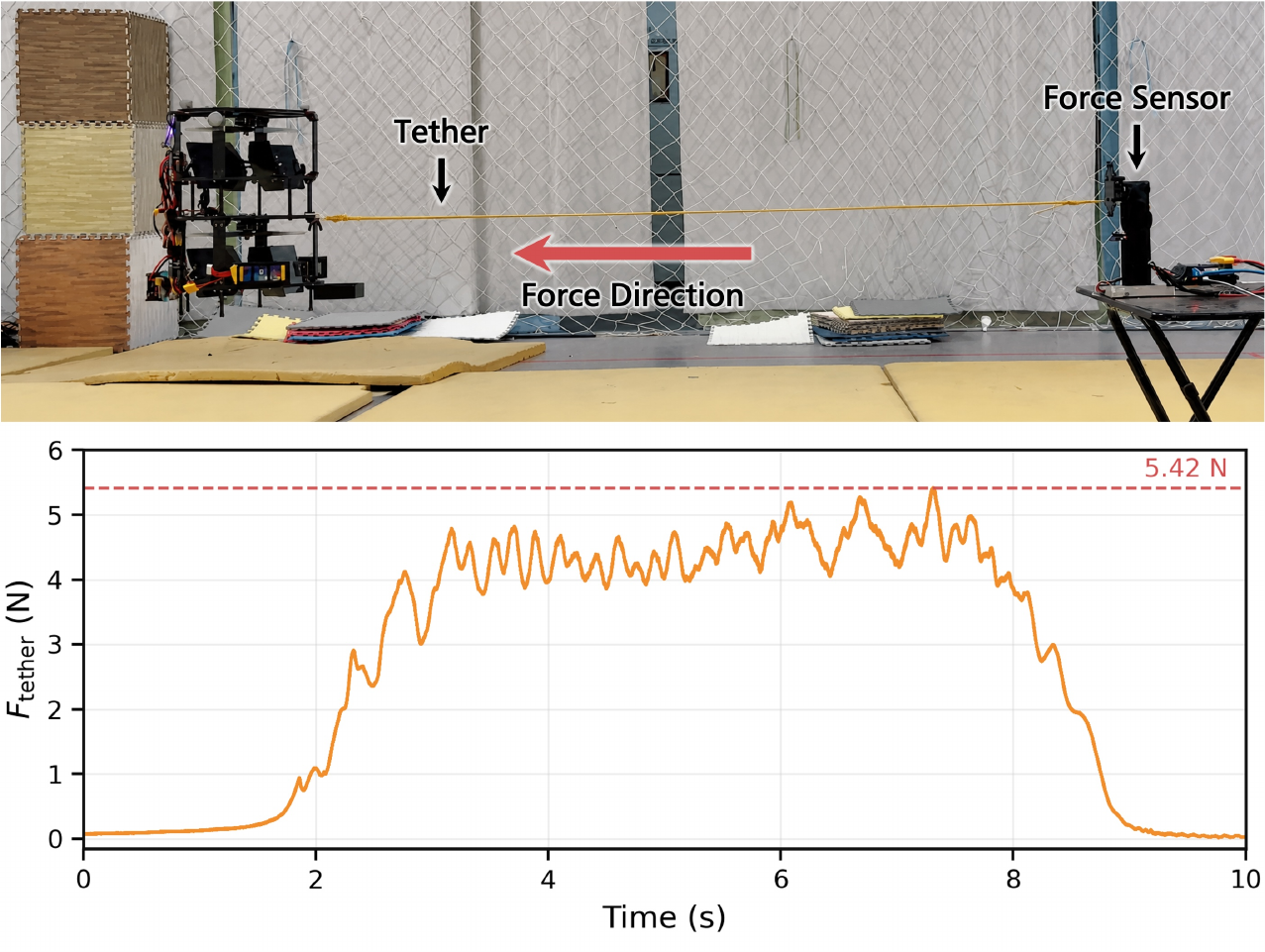}
    \caption{Tethered lateral-force characterization at zero roll and pitch commands. The upper panel shows the tether and force-sensor setup, and the lower panel shows the measured tether force.}
    \label{fig:lateral_force_exp}
\end{figure}

\section{Experiments}
This section experimentally validates the lateral force capability, the nonlinear aerodynamic allocation model, the $\mathcal{L}_1$ disturbance compensation, and the physical-interaction capability of the integrated system.

\subsection{Experimental Setup}
All experiments were conducted in an indoor motion-capture environment using the upgraded prototype described in Section~\ref{sec:prototype}. The onboard estimator fused IMU and motion-capture measurements, and the controller ran onboard at $100~\mathrm{Hz}$. When applicable, four controller configurations were compared: linear control allocation (LCA), $\mathcal{L}_1$ adaptive control with LCA (L1-LCA), nonlinear control allocation (NLCA), and the proposed method combining $\mathcal{L}_1$ adaptive control with NLCA. The same vehicle, actuator limits, and feedback gains were used within each comparative test.

\subsection{Lateral Force Characterization}
The lateral force capability was evaluated using a tethered force-measurement setup. FLOAT Drone hovered with zero roll and pitch commands while its position was gradually shifted to tension the tether. As shown in Fig.~\ref{fig:lateral_force_exp}, the maximum tether force reached $5.42~\mathrm{N}$. With the test mass of $1.85~\mathrm{kg}$, this corresponds to $30\%$ of the vehicle weight. An underactuated multirotor would require approximately $17^\circ$ of body tilt to generate the same lateral force, whereas the measured roll and pitch angles of FLOAT Drone remained below $2.3^\circ$.

\subsection{Validation of Modeling and Control Allocation}
\subsubsection{Three-Dimensional Trajectory Tracking}
\begin{figure}[t]
    \centering
    \includegraphics[width=0.8\linewidth]{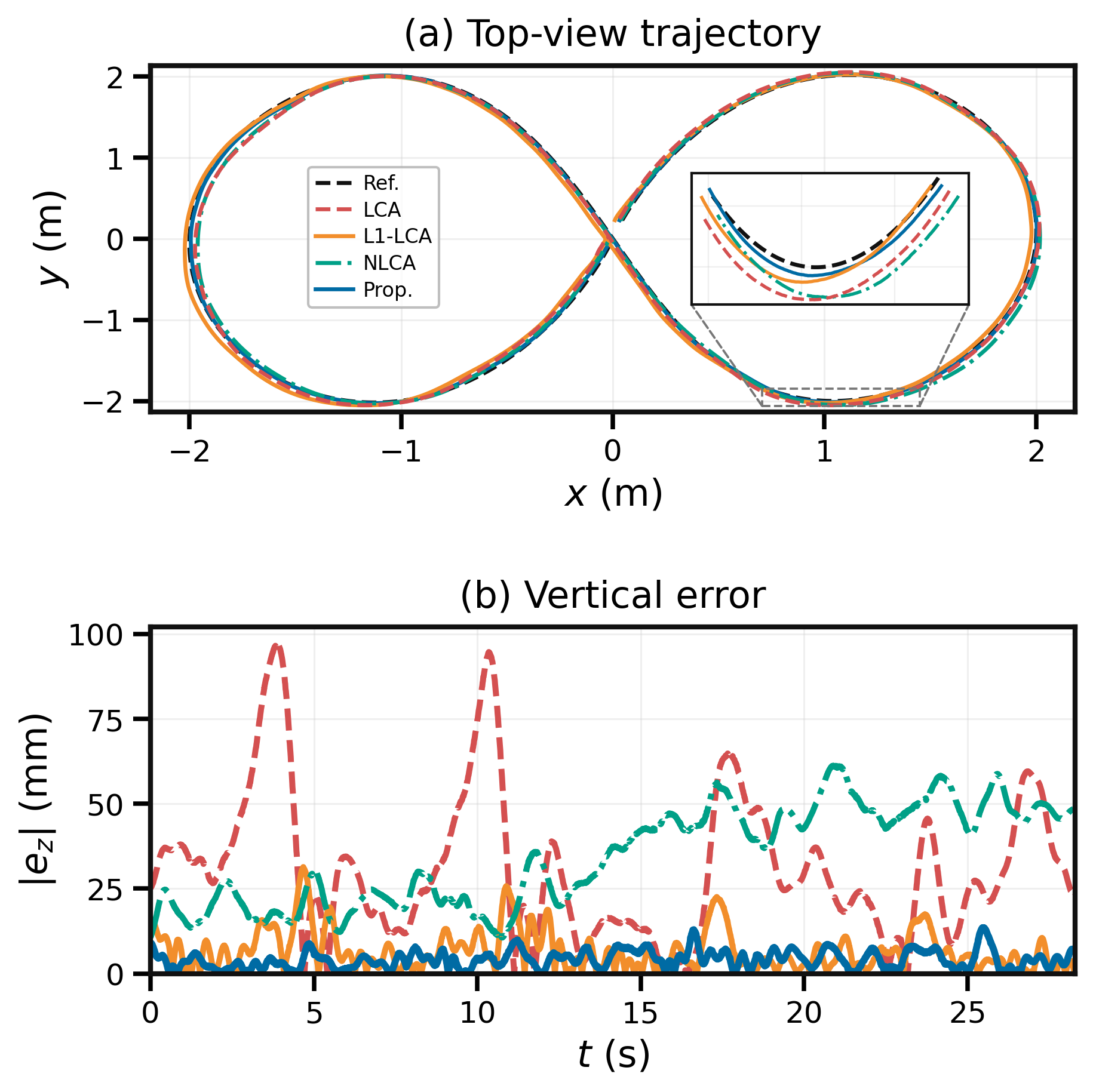}
    \caption{Three-dimensional trajectory tracking. (a) Top-view trajectory with zoomed inset. (b) Vertical tracking error during the same trajectory.}
    \label{fig:traj_tracking_exp}
\end{figure}

FLOAT Drone was commanded to track a spatial 8-shaped trajectory with coupled translational and yaw motion. The maximum translational speed and acceleration were $1~\mathrm{m/s}$ and $0.51~\mathrm{m/s^2}$, respectively. Fig.~\ref{fig:traj_tracking_exp} shows the top-view trajectory and vertical-error response, and Table~\ref{tab:tracking_performance} shows that $\mathcal{L}_1$ compensation substantially reduced vertical tracking error, and the proposed NLCA+$\mathcal{L}_1$ controller achieved the lowest position RMSE, attitude RMSE, and maximum $z$ error.

\begin{figure}[t]
    \centering
    \includegraphics[width=0.8\linewidth]{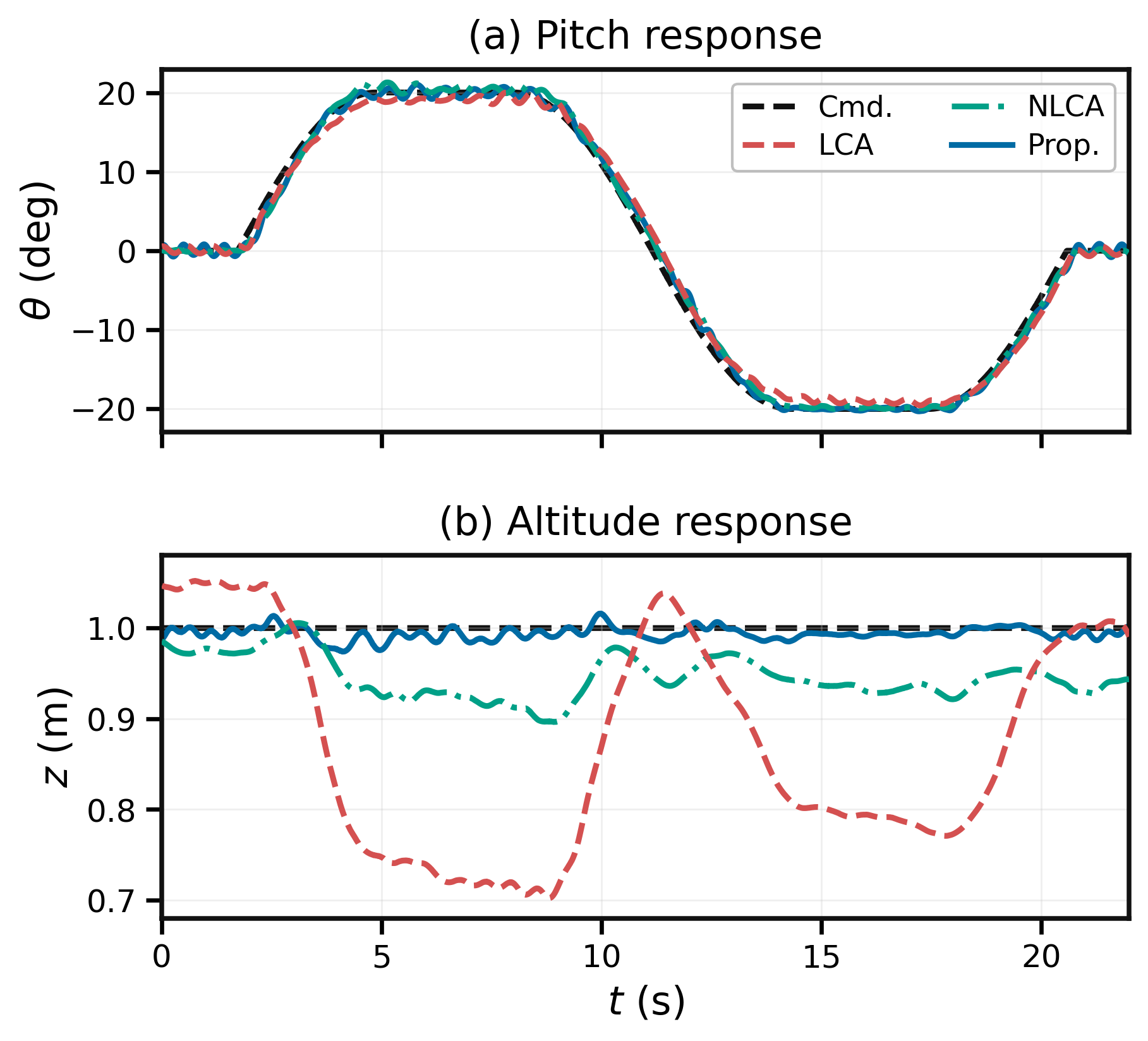}
    \caption{Hovering attitude-transition test. (a) Pitch response during the commanded $\pm20^\circ$ transition. (b) Altitude response during the same maneuver.}
    \label{fig:hover_attitude_exp}
\end{figure}

\begin{table}[t]
\centering
\caption{Trajectory Tracking and Attitude Transition Performance.}
\label{tab:tracking_performance}
\footnotesize
\setlength{\tabcolsep}{2.0pt}
\renewcommand{\arraystretch}{1.04}
\begin{tabular}{@{}>{\raggedright\arraybackslash}p{0.40\linewidth}@{\hspace{0.45em}}r@{\hspace{0.75em}}r@{\hspace{0.75em}}r@{}}
\toprule
Controller & \makecell[c]{Pos. RMSE\\(mm)} & \makecell[c]{Att. RMSE\\(deg)} & \makecell[c]{Max $|e_z|$\\(mm)} \\
\midrule
\emph{3-D trajectory tracking} & & & \\
LCA & 49.7 & 3.39 & 97.3 \\
L1-LCA & 16.2 & 2.97 & 31.4 \\
NLCA & 46.7 & 3.11 & 61.2 \\
\textbf{Proposed} & \textbf{13.8} & \textbf{2.91} & \textbf{13.5} \\
\midrule
\emph{Hovering attitude transition} & & & \\
LCA & 180.2 & 1.40 & 297.6 \\
NLCA & 65.4 & 1.07 & 103.6 \\
\textbf{Proposed} & \textbf{11.0} & \textbf{1.00} & \textbf{26.0} \\
\bottomrule
\end{tabular}
\end{table}

\subsubsection{Hovering Attitude Transition}
To test large-attitude fully actuated flight, FLOAT Drone was commanded to hold position while the desired pitch angle switched between $+20^\circ$ and $-20^\circ$. As shown in Fig.~\ref{fig:hover_attitude_exp} and Table~\ref{tab:tracking_performance}, all controllers tracked the pitch command, but NLCA greatly reduced the altitude deviation caused by large attitude transitions, and the proposed controller further suppressed the residual steady-state error.

\subsection{Robustness Under Disturbances and Uncertainties}
\subsubsection{Ground-Effect Disturbance Test}
\begin{figure}[t]
    \centering
    \includegraphics[width=0.9\linewidth]{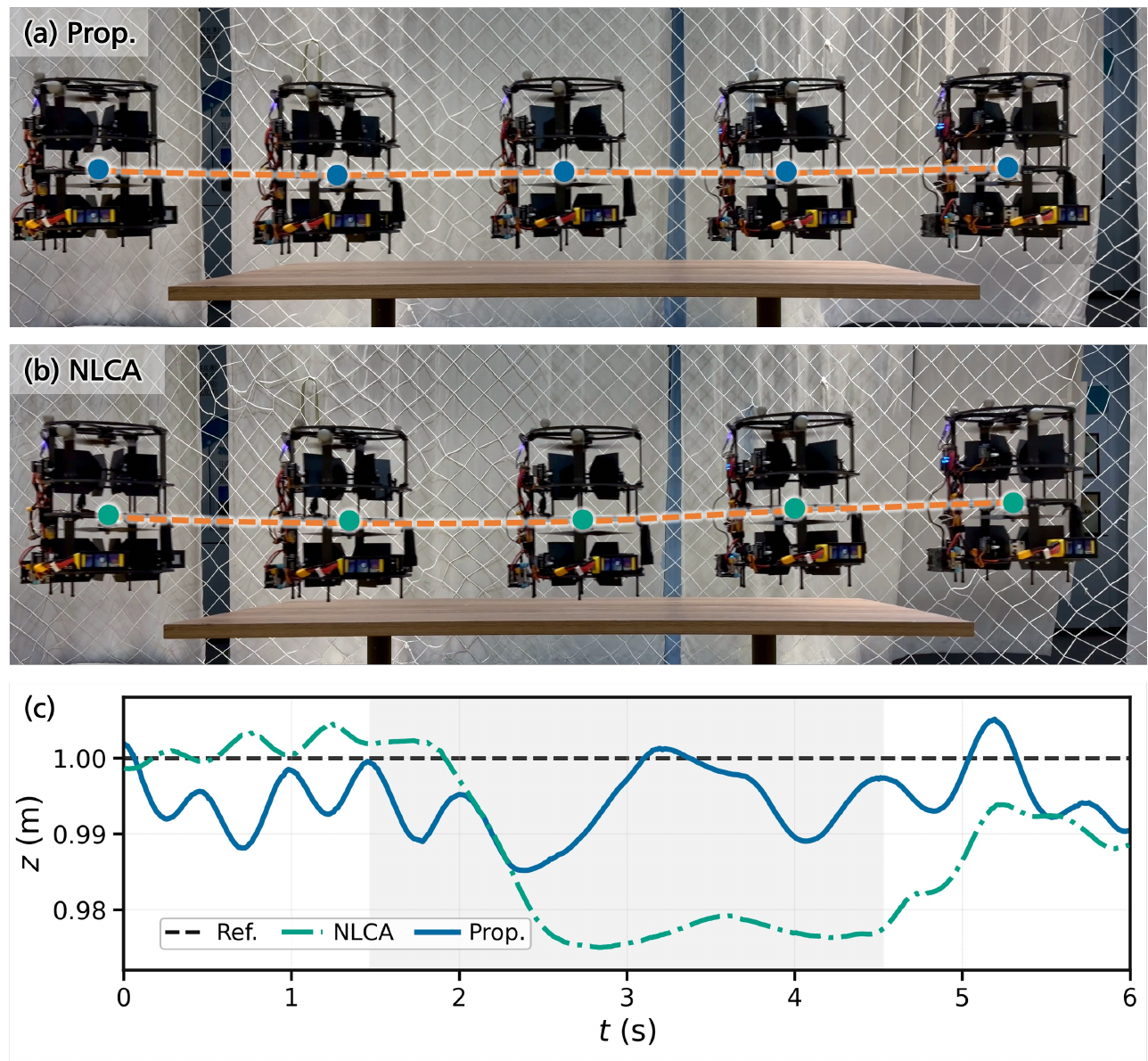}
    \caption{Ground-effect disturbance test. (a) Overflight sequence with the proposed controller. (b) Overflight sequence with NLCA. (c) Altitude response; the shaded region denotes the table-overflight interval.}
    \label{fig:ground_effect_exp}
\end{figure}

The proposed method was next compared with NLCA alone in a table-overflight test at $0.5~\mathrm{m/s}$. The commanded vehicle-center height was $1.0~\mathrm{m}$, leaving only about $50~\mathrm{mm}$ clearance between the vehicle bottom and the tabletop and thereby inducing a strong ground-effect disturbance. As shown in Fig.~\ref{fig:ground_effect_exp}, the proposed method reduced the maximum $z$ error in the table region from $25.2~\mathrm{mm}$ to $15.0~\mathrm{mm}$, shortened the recovery time from $2.61~\mathrm{s}$ to $1.02~\mathrm{s}$, and reduced the steady $z$ error from $-23.0~\mathrm{mm}$ to $-4.27~\mathrm{mm}$.

\subsubsection{Sudden Payload Attachment Test}
\begin{figure}[t]
    \centering
    \includegraphics[width=1.0\linewidth]{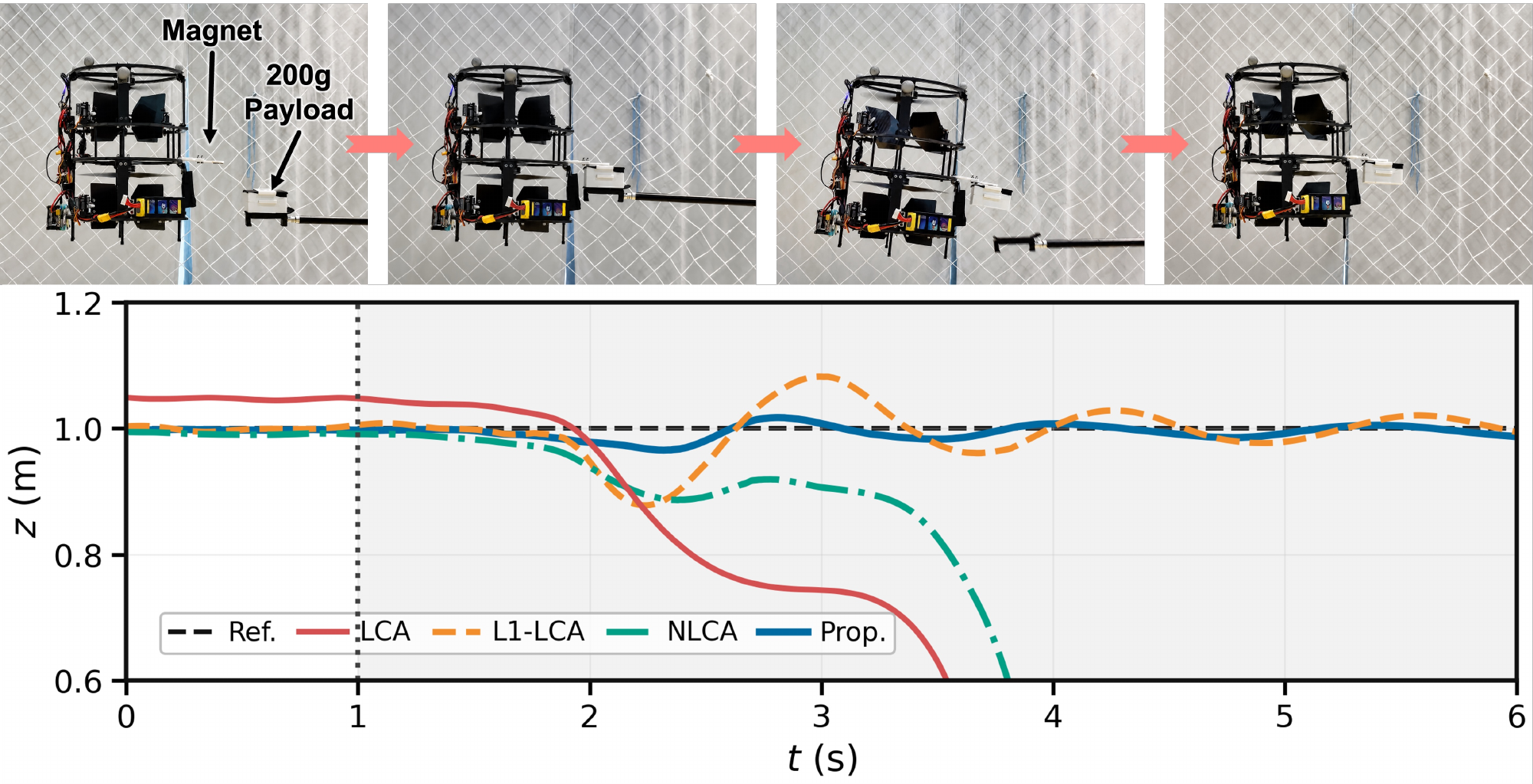}
    \caption{Representative $200~\mathrm{g}$ sudden-payload response. The snapshots illustrate the magnetic payload-attachment process. The vertical line marks the attachment instant, and the shaded region denotes the post-attachment interval.}
    \label{fig:payload_200g}
\end{figure}

Robustness to abrupt load changes was evaluated by magnetically attaching $100~\mathrm{g}$ and $200~\mathrm{g}$ payloads to the hovering vehicle. Table~\ref{tab:payload_results} reports the post-attachment performance. For the $100~\mathrm{g}$ payload, all configurations completed the test, but the two $\mathcal{L}_1$-based controllers produced much lower position and attitude errors. For the $200~\mathrm{g}$ payload, LCA and NLCA without adaptive compensation failed, whereas L1-LCA and the proposed method maintained stable flight. The representative $200~\mathrm{g}$ response in Fig.~\ref{fig:payload_200g} illustrates this recovery behavior. The proposed method reduced the maximum $z$ error from $122.0~\mathrm{mm}$ to $34.7~\mathrm{mm}$ relative to L1-LCA, indicating that adaptive disturbance rejection and nonlinear wrench allocation are complementary.

\begin{table}[t]
\centering
\caption{Sudden Payload Attachment Performance.}
\label{tab:payload_results}
\footnotesize
\setlength{\tabcolsep}{1.8pt}
\renewcommand{\arraystretch}{1.04}
\begin{tabular}{@{}>{\raggedright\arraybackslash}p{0.20\linewidth}@{\hspace{0.35em}}l@{\hspace{0.45em}}r@{\hspace{0.45em}}r@{\hspace{0.45em}}r@{\hspace{0.45em}}r@{}}
\toprule
Controller & Status & \makecell[c]{Pos. RMSE\\(mm)} & \makecell[c]{$z$ RMSE\\(mm)} & \makecell[c]{Max $|e_z|$\\(mm)} & \makecell[c]{Att. RMSE\\(deg)} \\
\midrule
\emph{$100~\mathrm{g}$ payload} & & & & & \\
LCA & Success & 35.9 & 24.9 & 76.9 & 2.68 \\
L1-LCA & Success & 8.85 & 7.48 & 32.4 & 0.743 \\
NLCA & Success & 50.4 & 45.9 & 53.3 & 1.97 \\
\textbf{Proposed} & \textbf{Success} & \textbf{8.23} & \textbf{6.40} & \textbf{18.9} & \textbf{0.557} \\
\midrule
\emph{$200~\mathrm{g}$ payload} & & & & & \\
LCA & Failure & -- & -- & -- & -- \\
L1-LCA & Success & 38.1 & 36.5 & 122.0 & 2.96 \\
NLCA & Failure & -- & -- & -- & -- \\
\textbf{Proposed} & \textbf{Success} & \textbf{13.4} & \textbf{11.9} & \textbf{34.7} & \textbf{1.59} \\
\bottomrule
\end{tabular}
\end{table}

\subsubsection{Steady-State Error Rejection in Long-Duration Hovering}
\begin{figure}[t]
    \centering
    \includegraphics[width=1.0\linewidth]{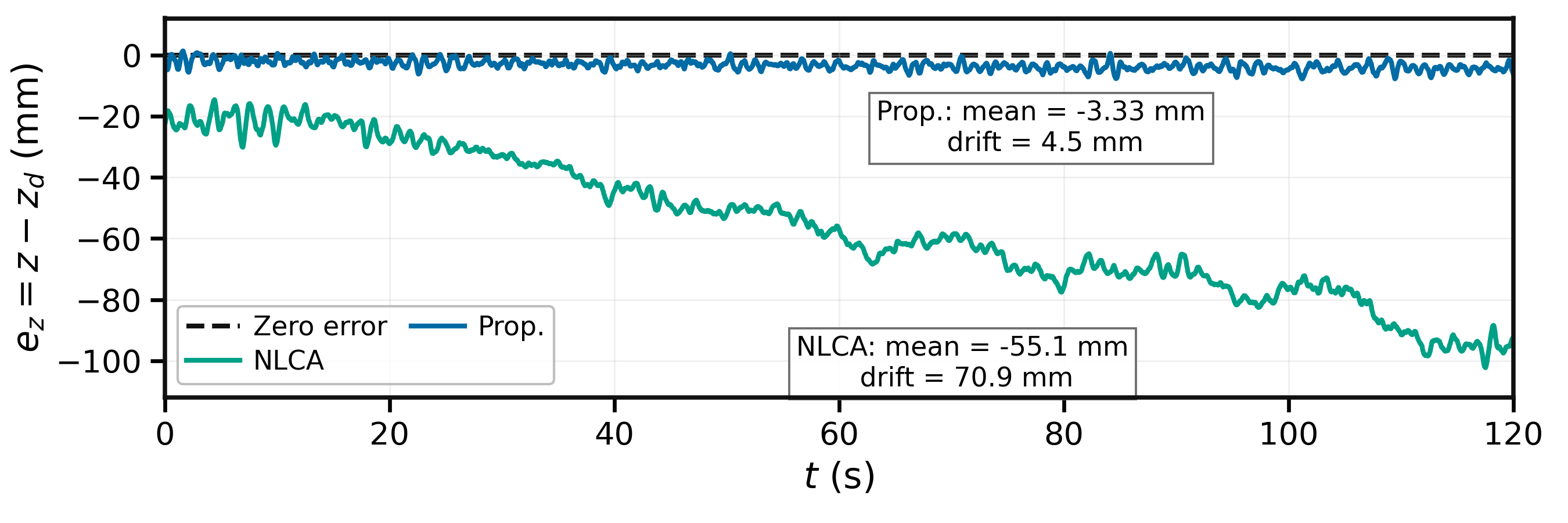}
    \caption{Steady-state altitude-error rejection in long-duration hovering, where $e_z=z-z_d$.}
    \label{fig:long_duration_hovering}
\end{figure}

A two-minute hovering segment was used to evaluate steady-state error rejection under slowly varying actuator-model mismatch. As shown in Fig.~\ref{fig:long_duration_hovering}, the proposed controller reduced the mean altitude error from $-55.1~\mathrm{mm}$ to $-3.33~\mathrm{mm}$ and substantially suppressed long-term drift without degrading hover stability.

\subsection{Drawer Manipulation Demonstration}
Finally, FLOAT Drone was tested in a drawer push--pull manipulation task using a side-mounted 3-D printed hook, as shown in Fig.~\ref{fig:drawer_manipulation}. The clearance between the handle and the front panel was only $2~\mathrm{cm}$, requiring accurate insertion and close-proximity positioning. The vehicle inserted the hook, pulled the drawer open, and pushed it closed while keeping zero roll and pitch commands. During the full task, the maximum attitude deviation was $3.41^\circ$, and the off-axis tracking error remained within $12.2~\mathrm{mm}$. These results show that FLOAT Drone maintained close-proximity positioning accuracy during bidirectional drawer manipulation through the $2~\mathrm{cm}$ handle clearance without requiring large body tilt.

\section{Conclusion}

This paper presented FLOAT Drone, a fully actuated coaxial UAV that uses servo-driven control surfaces for 6-DoF wrench generation with reduced target-facing lateral airflow. Force-matched CFD, precision wrench modeling, nonlinear allocation, and geometric $\mathcal{L}_1$ adaptive control verified its airflow advantage, improved tracking and attitude transitions, disturbance robustness, and close-proximity drawer push--pull capability through a $2~\mathrm{cm}$ handle clearance.

Overall, this work offers the aerial manipulation community a complete system-level solution for close-proximity physical interaction. A remaining limitation is that the demonstrated manipulation relies on a simple side-mounted hook and controlled indoor conditions rather than a general-purpose end-effector and autonomous task execution. Future work will integrate lightweight modular end-effectors with the vehicle, co-design their mass and mounting geometry with the available wrench authority, and validate target-localized insertion, pulling, pushing, and grasping tasks beyond motion-capture-based experiments.


\bibliographystyle{IEEEtran}
\bibliography{bib}

\makeatletter
\def\@IEEEBIOskipN{2\baselineskip}
\makeatother

\begin{IEEEbiography}[{\includegraphics[width=1in,height=1.25in,clip,keepaspectratio]{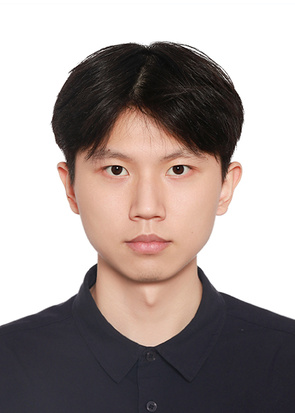}}]{Junxiao Lin}
received the M.S. degree in control engineering from Zhejiang University, Hangzhou, China, in 2026.

His research interests include robot system design and motion control.
\end{IEEEbiography}

\begin{IEEEbiography}[{\includegraphics[width=1in,height=1.25in,clip,keepaspectratio]{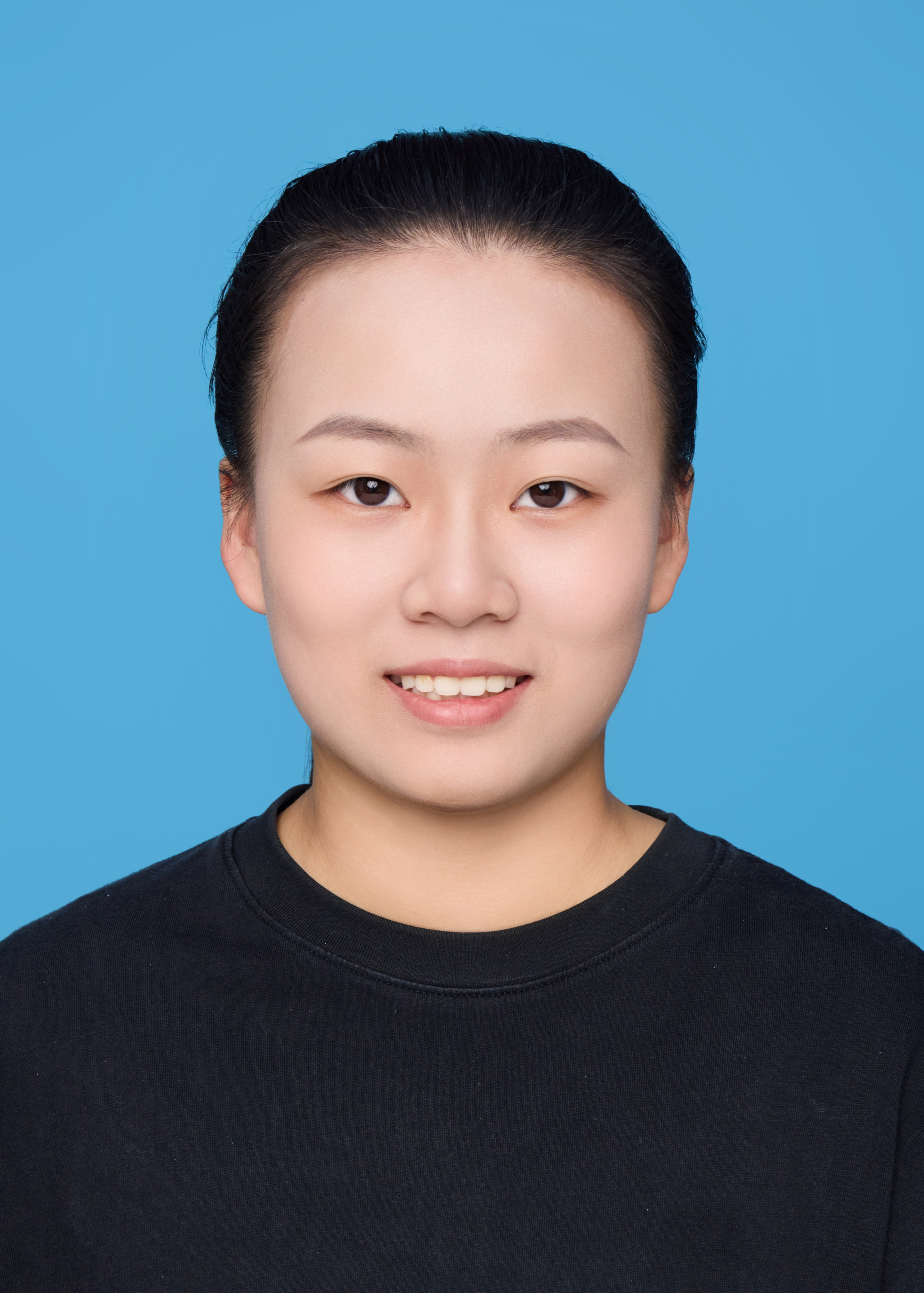}}]{Kehan Zhou}
is currently pursuing the B.Eng. degree in automation with the College of Control Science and Engineering and Chu Kochen Honors College, Zhejiang University, Hangzhou, China. She is currently an undergraduate researcher with the FAST Lab under the supervision of Prof. Fei Gao. 

Her research interests include mobile manipulation, robot control, and novel robotic configurations.
\end{IEEEbiography}

\begin{IEEEbiography}[{\includegraphics[width=1in,height=1.25in,clip,keepaspectratio]{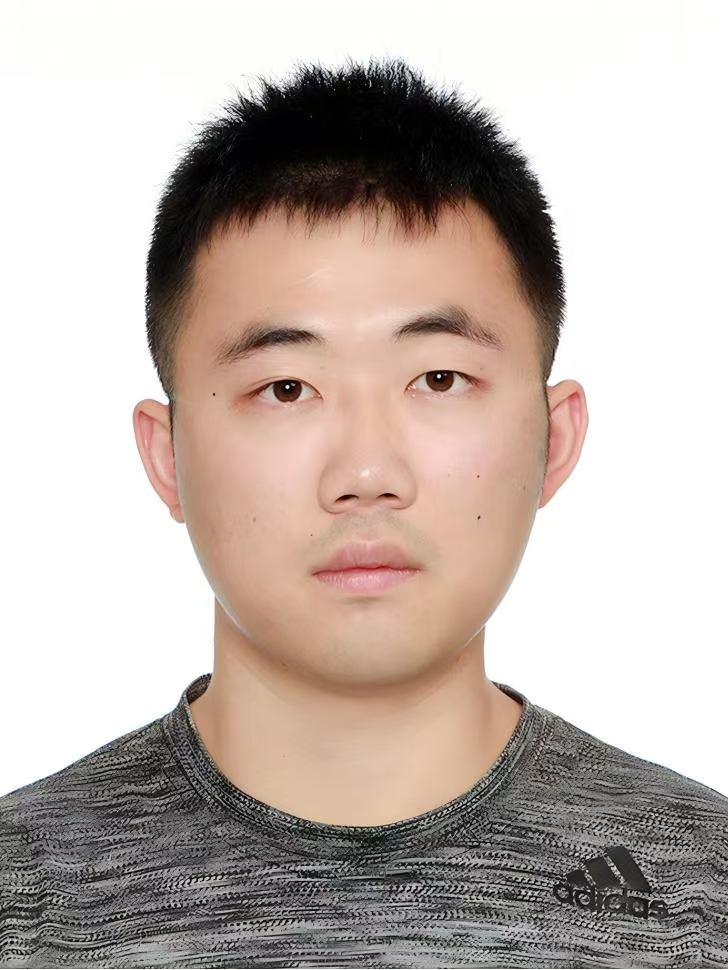}}]{Shuhang Ji}
received the B.Eng. degree in control science and engineering from Zhejiang University, Hangzhou, China, in 2025, currently working as a robotics system engineer in Differential Robotics (Hangzhou) Technology Co., Ltd.

His research interests include aerial manipulator, robot system and control.
\end{IEEEbiography}

\begin{IEEEbiography}[{\includegraphics[width=1in,height=1.25in,clip,keepaspectratio]{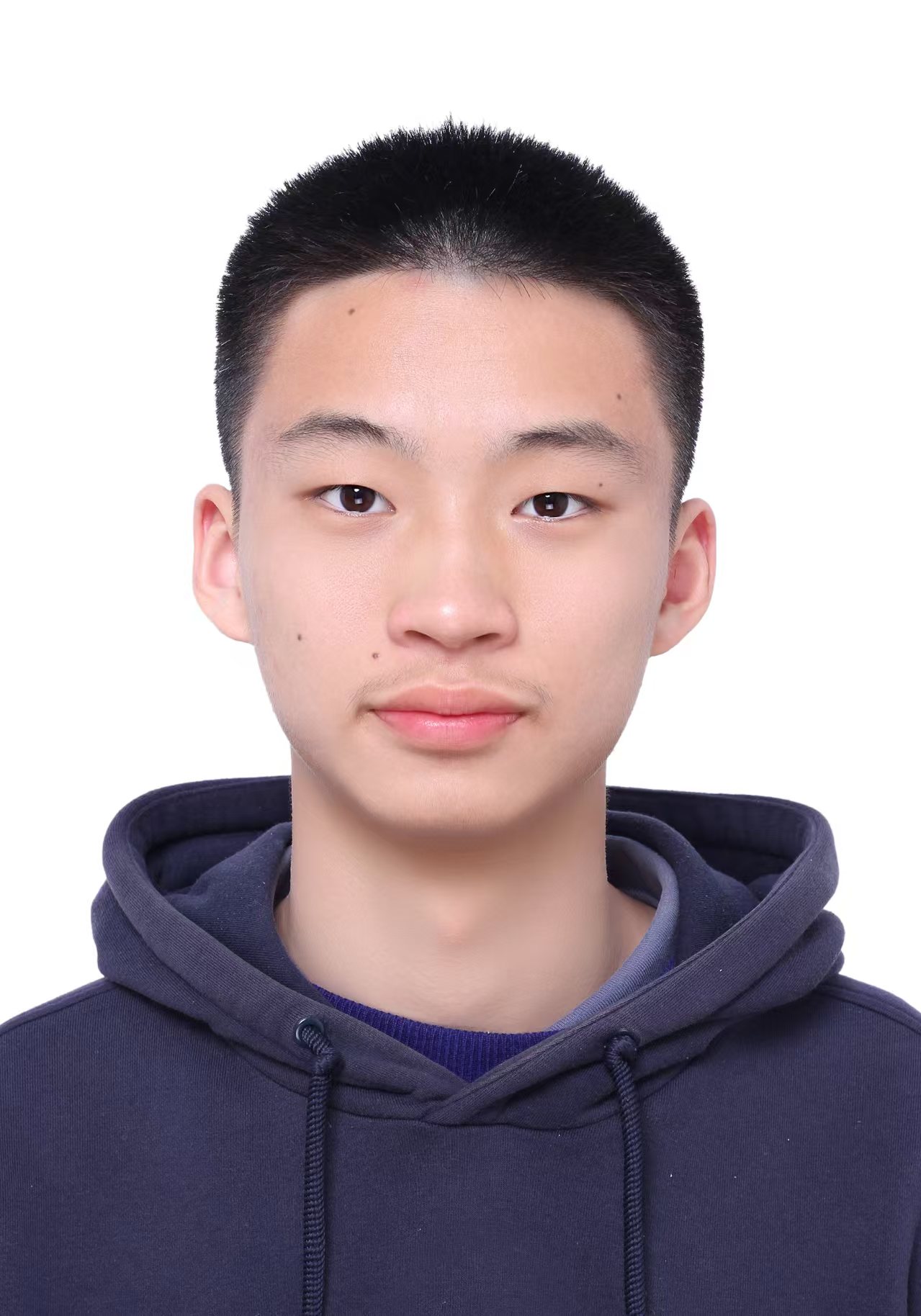}}]{Yimin Peng}
is currently working toward the bachelor's degree in automation from Zhejiang University, Hangzhou, China. He is also currently a research intern with the FAST Lab, Zhejiang University.

His research interests include autonomous aerial robotics, motion planning, and reinforcement learning for robotics.
\end{IEEEbiography}

\begin{IEEEbiography}[{\includegraphics[width=1in,height=1.25in,clip,keepaspectratio]{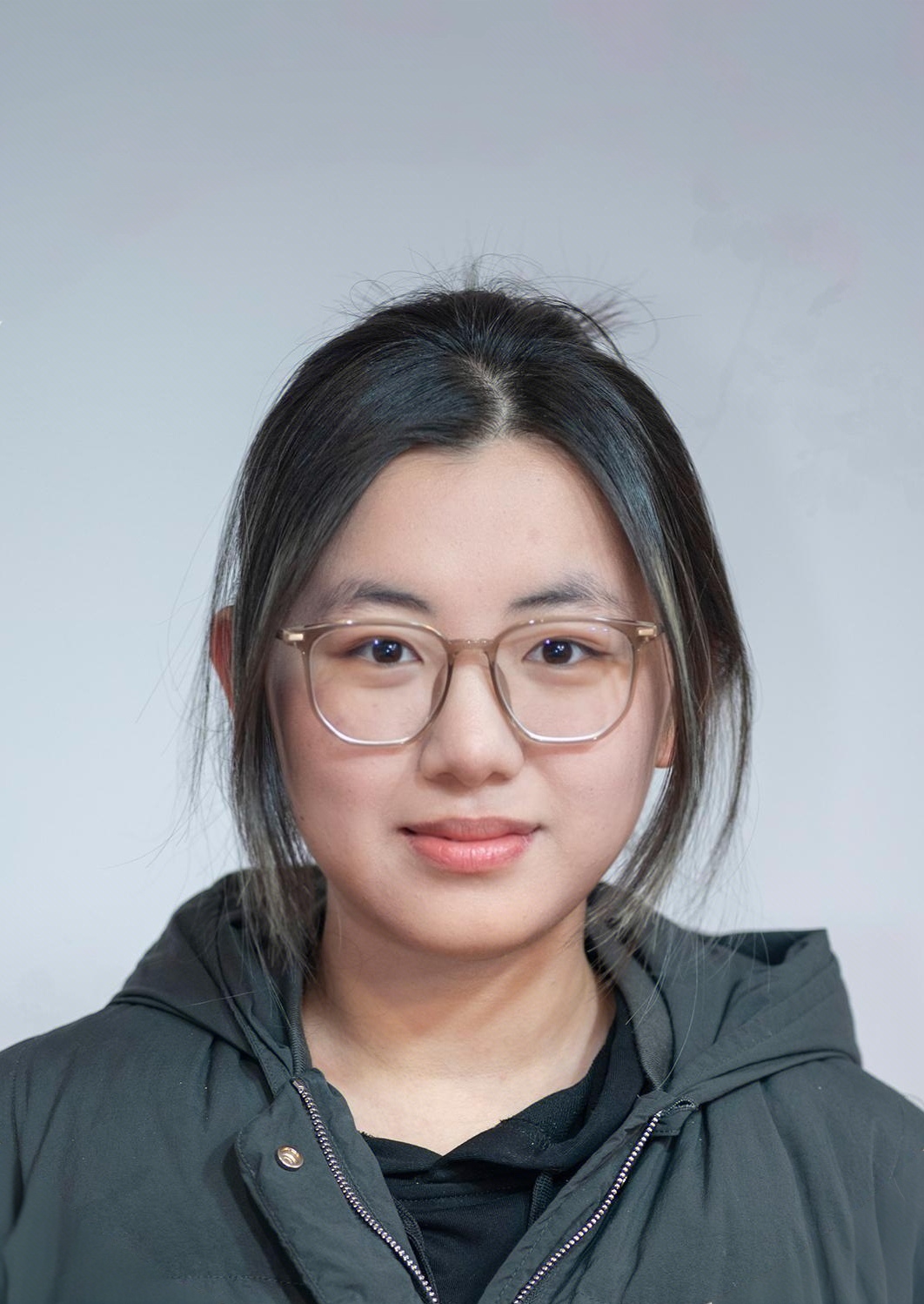}}]{Shen Wang}
is currently pursuing the bachelor's degree in automation at the College of Control Science and Engineering, Zhejiang University, Hangzhou, China. She is also pursuing a minor in the Advanced Honor Class of Engineering Education (ACEE) at Chu Kochen Honors College, Zhejiang University.

Her research interests include motion planning, control, and machine learning.
\end{IEEEbiography}

\begin{IEEEbiography}[{\includegraphics[width=1in,height=1.25in,clip,keepaspectratio]{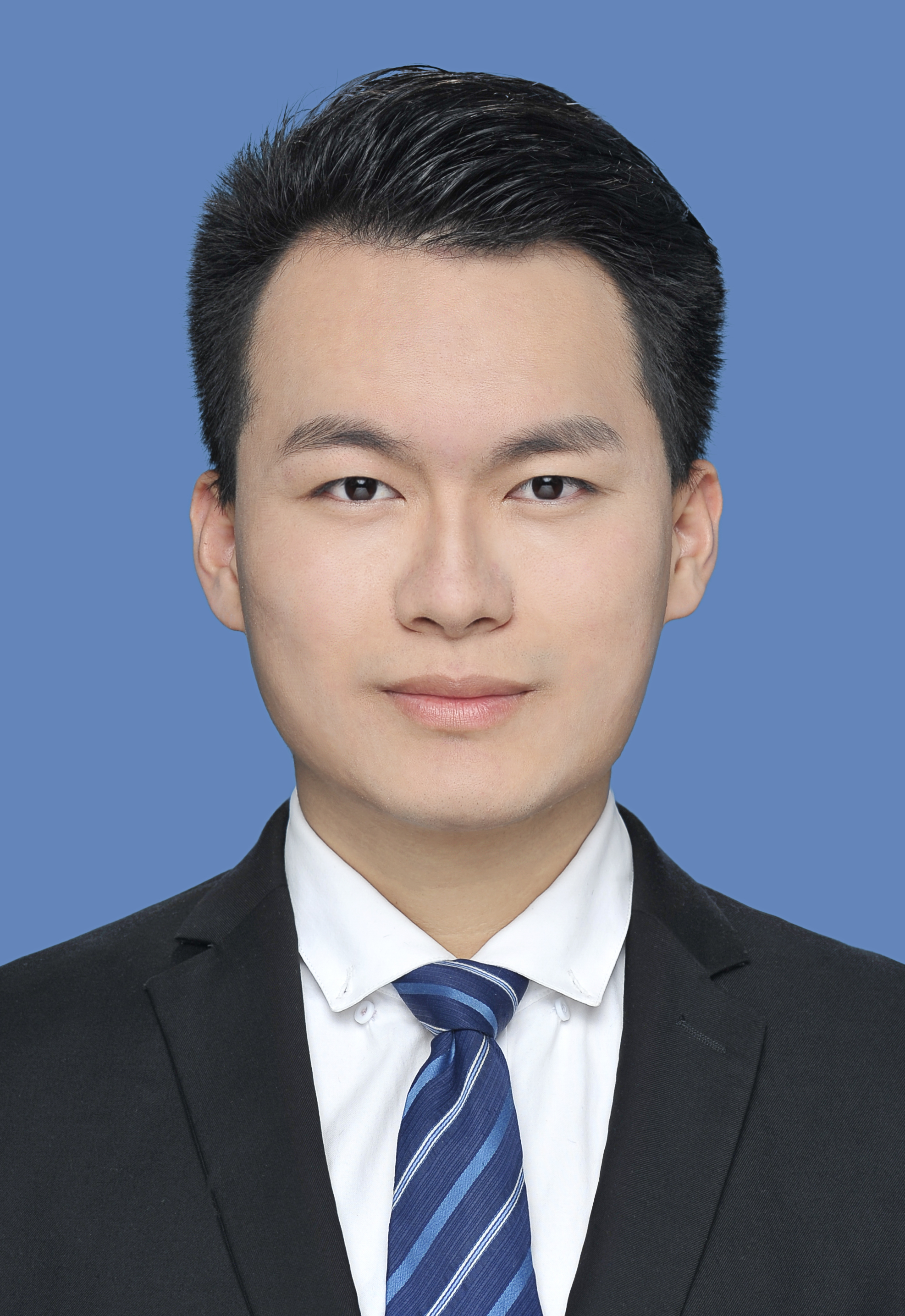}}]{Jialiang Hou}
received the Ph.D. degree in computer application technology from Fudan University, Shanghai, China, in 2025.
He is currently a assistant research fellow/postdoctoral fellowship in the College of Control Science and Engineering, Zhejiang University, Hangzhou, China, supervised by Professor Fei Gao.

His research interests include autonomous navigation, swarm intelligence, embodied intelligence for mobile robots.
\end{IEEEbiography}

\begin{IEEEbiography}[{\includegraphics[width=1in,height=1.25in,clip,keepaspectratio]{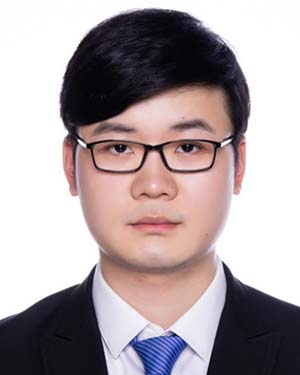}}]{Fei Gao}
(Member, IEEE) received the Ph.D. degree in electronic and computer engineering from the Hong Kong University of Science and Technology, Hong Kong, in 2019.

He is currently a Tenured Associate Professor with the Department of Control Science and
Engineering, Zhejiang University, Hangzhou, China. He is also the Founder of Differential Robotics Ltd. His research interests include aerial robots, autonomous navigation,
motion planning, optimization, and localization and mapping.\end{IEEEbiography}

\end{document}